%% file: 0_main.tex
\pdfoutput=1
\documentclass[11pt]{article}
\usepackage[]{ACL2023}
\usepackage{times}
\usepackage{latexsym}
\usepackage[T1]{fontenc}
\usepackage[utf8]{inputenc}
\usepackage{microtype}
\usepackage{inconsolata}

\usepackage{booktabs}
\usepackage{graphicx}
\usepackage{xspace}
\usepackage{xparse}

\usepackage{xcolor,colortbl}
\usepackage{collcell}
\usepackage{xstring}
\usepackage{textcomp}

\usepackage{color}
\usepackage{soul}
\usepackage{tikz}

\usepackage{arydshln}
\usepackage{multirow}

\usepackage{censor}

\NewDocumentCommand{\DIV}{om}{%
  \IfValueT{#1}{\setcounter{#2}{\numexpr#1-1\relax}}%
  \csname #2\endcsname
}

\makeatletter
\def\SOUL@hlpreamble{%
    \setul{}{3.25ex}%
    \let\SOUL@stcolor\SOUL@hlcolor
    \dimen@\SOUL@ulthickness
    \dimen@i=-.75ex %
    \advance\dimen@i-.5\dimen@
    \edef\SOUL@uldepth{\the\dimen@i}%
    \let\SOUL@ulcolor\SOUL@stcolor
    \SOUL@ulpreamble
}
\makeatother

\usepackage{array}
\makeatletter
\newcommand{\thickhline}{%
    \noalign {\ifnum 0=`}\fi \hrule height 1pt
    \futurelet \reserved@a \@xhline
}
\newcolumntype{"}{@{\hskip\tabcolsep\vrule width 1pt\hskip\tabcolsep}}

\newcolumntype{|}{@{\hskip\tabcolsep\vrule width 0.5pt\hskip\tabcolsep}}

\makeatother

\newcommand*{\MinNumber}{-1}%
\newcommand*{\MidNumber}{0} %
\newcommand*{\MaxNumber}{1}%

\definecolor{accgood}{HTML}{4caf50}
\definecolor{accneutral}{HTML}{ffeb3b}
\definecolor{accbad}{HTML}{f44336}

\def\acc#1{
    \ifdim#1pt<\MidNumber pt
        \pgfmathsetmacro{\PercentColor}{100*((\MidNumber - #1)/(\MidNumber - \MinNumber))}
        \xdef\PercentColorr{\PercentColor}
        {\setlength{\fboxsep}{2pt}\tcbox[on line,colframe=accbad!\PercentColorr!accneutral!60,boxsep=0pt,left=2pt,right=2pt, enlarge top by=0.005cm, enlarge bottom by=0.005cm, arc=0.5pt, top=1pt,bottom=1pt,colback=accbad!\PercentColorr!accneutral!60]{\scriptsize #1}}
    \else
        \pgfmathsetmacro{\PercentColor}{100*((#1 - \MidNumber)/(\MaxNumber - \MidNumber))} %
        \xdef\PercentColorr{\PercentColor}
        {\setlength{\fboxsep}{2pt}\tcbox[on line,colframe=accgood!\PercentColorr!accneutral!60,boxsep=0pt,left=2pt,right=2pt, enlarge top by=0.005cm, enlarge bottom by=0.005cm, arc=0.5pt, top=1pt,bottom=1pt,colback=accgood!\PercentColorr!accneutral!60]{\scriptsize \phantom{-}#1}}
    \fi
}

\definecolor{toxgood}{HTML}{f44336}
\definecolor{toxneutral}{HTML}{ffeb3b}
\definecolor{toxbad}{HTML}{4caf50}

\def\tox#1{
    \ifdim#1pt<\MidNumber pt
        \pgfmathsetmacro{\PercentColor}{100*((\MidNumber - #1)/(\MidNumber - \MinNumber))}
        \xdef\PercentColorr{\PercentColor}
        {\setlength{\fboxsep}{2pt}\tcbox[on line,colframe=toxbad!\PercentColorr!toxneutral!60,boxsep=0pt,left=2pt,right=2pt, enlarge top by=0.005cm, enlarge bottom by=0.005cm, arc=0.5pt, top=1pt,bottom=1pt,colback=toxbad!\PercentColorr!toxneutral!60]{\scriptsize #1}}
    \else
        \pgfmathsetmacro{\PercentColor}{100*((#1 - \MidNumber)/(\MaxNumber - \MidNumber))} %
        \xdef\PercentColorr{\PercentColor}
        {\setlength{\fboxsep}{2pt}\tcbox[on line,colframe=toxgood!\PercentColorr!toxneutral!60,boxsep=0pt,left=2pt,right=2pt, enlarge top by=0.005cm, enlarge bottom by=0.005cm, arc=0.5pt, top=1pt,bottom=1pt,colback=toxgood!\PercentColorr!toxneutral!60]{\scriptsize \phantom{-}#1}}
    \fi
}

\definecolor{vargood}{HTML}{1E88E5}
\definecolor{varneutral}{HTML}{90CAF9}
\definecolor{varbad}{HTML}{E3F2FD}

\newcommand*{\MinMRR}{0}%
\newcommand*{\MidMRR}{1} %
\newcommand*{\MaxMRR}{5}%

\def\varz#1{
    \ifdim#1pt<\MidMRR pt
        \pgfmathsetmacro{\PercentColor}{100*((\MidMRR - #1)/(\MidMRR - \MinMRR))}
        \xdef\PercentColorr{\PercentColor}
        {\setlength{\fboxsep}{2pt}\tcbox[on line,colframe=varbad!\PercentColorr!varneutral!60,boxsep=0pt,left=2pt,right=2pt, enlarge top by=0.005cm, enlarge bottom by=0.005cm, arc=0.5pt, top=1pt,bottom=1pt,colback=varbad!\PercentColorr!varneutral!60]{\scriptsize #1}}
    \else
        \pgfmathsetmacro{\PercentColor}{100*((#1 - \MidMRR)/(\MaxMRR - \MidMRR))} %
        \xdef\PercentColorr{\PercentColor}
        {\setlength{\fboxsep}{2pt}\tcbox[on line,colframe=vargood!\PercentColorr!varneutral!60,boxsep=0pt,left=2pt,right=2pt, enlarge top by=0.005cm, enlarge bottom by=0.005cm, arc=0.5pt, top=1pt,bottom=1pt,colback=vargood!\PercentColorr!varneutral!60]{\scriptsize #1}}
    \fi
}

\newcommand{\framework}{{\fontfamily{qtm}\selectfont\textsc{NLPositionality}}\xspace}

\newcommand{\custo}{\small\fontfamily{qcs}\selectfont}

\newcommand{\dataseticon}[2]{{\custo\includegraphics[height=1.3\fontcharht\font`\B]{#1}{\kern 0.15em}#2}\xspace}

\newcommand{\newText}[1]{{\textcolor{black}{#1}}}

\usepackage{colortbl}
\usepackage{tcolorbox}
\definecolor{bg}{RGB}{248,248,248}
\definecolor{bgbord}{RGB}{235,235,235}
\definecolor{extitle}{HTML}{000000}
\newtcolorbox{examplebox}[1][]{
    colback=bg,colframe=bgbord,arc=3pt,left=3pt, top=3pt, bottom=3pt, right=3pt,
    width=\linewidth, before=\par\smallskip\centering, IfValueTF={#1}{#1}{}
}

\title{
\framework: \\ Characterizing Design Biases of Datasets and Models}

\newcommand{\affuw}{$^{\dagger}$}
\newcommand{\affcmu}{$^{\ddagger}$}
\newcommand{\affai}{$^{\diamond}$}

\newcommand{\redact}[1]{#1}

\author{
Sebastin Santy\affuw\footnotemark[1] \quad
Jenny T. Liang\affcmu\footnotemark[1]\\
\textbf{Ronan Le Bras}\affai \quad
\textbf{Katharina Reinecke}\affuw \quad
\textbf{Maarten Sap}\affcmu\affai\\
\affuw University of Washington \quad
\affcmu Carnegie Mellon University \\
\affai Allen Institute for AI \\
\texttt{\small \{ssanty,reinecke\}@cs.washington.edu,}\\
\texttt{\small 
\{jtliang,maartensap\}@cs.cmu.edu, ronanlb@allenai.org} 
\\
}

\begin{document}
\maketitle

\begin{abstract}
\input{sections/00_abstract.tex}
\end{abstract}

\input{sections/01_introduction.tex}
\input{sections/02_design_bias.tex}
\input{sections/03_setup.tex}
\input{sections/04_case_studies.tex}
\input{sections/05_discussion.tex}
\input{sections/06_conclusion.tex}
\input{sections/98_ethics.tex}

\bibliography{custom}
\bibliographystyle{acl_natbib}

\input{sections/99_appendix.tex}

\end{document}

%% file: sections/00_abstract.tex
Design biases in NLP systems, such as performance differences for different populations, often stem from their creator's \textit{positionality}, i.e., views and lived experiences shaped by identity and background.
Despite the prevalence and risks of design biases, they are hard to quantify because researcher, system, and dataset positionality is often unobserved.
We introduce \framework, a framework for characterizing design biases and quantifying the positionality of NLP datasets and models.
Our framework continuously collects annotations from a diverse pool of volunteer participants on LabintheWild, and statistically quantifies alignment with dataset labels and model predictions.
We apply \framework to existing datasets and models for two tasks---social acceptability and hate speech detection. To date, we have collected $16,299$ annotations in over a year for $600$ instances from $1,096$ annotators across $87$ countries.
We find that datasets and models align predominantly with Western, White, college-educated, and younger populations. Additionally, certain groups, such as non-binary people and non-native English speakers, are further marginalized by datasets and models as they rank least in alignment across all tasks.
Finally, we draw from prior literature to discuss how researchers can examine their own positionality and that of their datasets and models, opening the door for more inclusive NLP systems.
\let\thefootnote\relax\footnotetext{\redact{* Equal contribution; work done while at the} \redact{Allen Institute for AI}}

%% file: sections/01_introduction.tex
\section{Introduction}
\begin{quote}
``Treating different things the same can generate as much inequality as treating the same things differently.'' 
\vspace{-1em}
\begin{flushright} -- \textit{Kimberlé Crenshaw} \end{flushright}
\vspace{-1.2em}
\end{quote}

\begin{figure}[t!]
    \centering
    \includegraphics[width=\columnwidth,page=1]{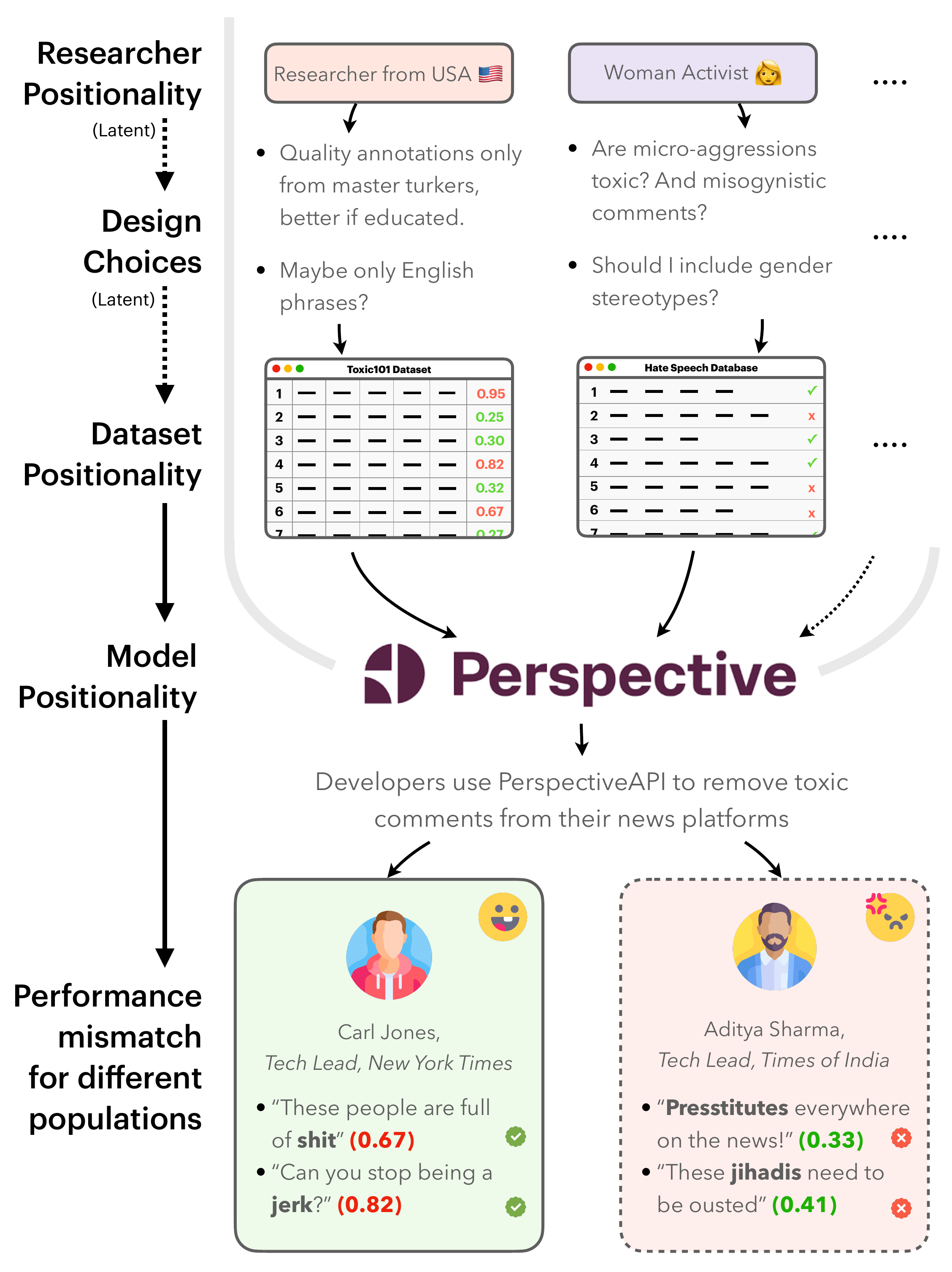}
    \caption{\textbf{Example Scenario.} Carl from the U.S. and Aditya from India both want to use Perspective API, but it works better for Carl than it does for Aditya. This is because toxicity researchers' positionalities lead them to make design choices that make toxicity datasets, and thus Perspective API, to have positionalities that are Western-centric.}
    \label{fig:positionality}
\end{figure}

When creating NLP datasets and models, researchers' design choices are partly influenced by their \textit{positionality}, i.e., their views shaped by their lived experience, identity, culture, and background~\cite{savin2013qualititative}.
While researcher positionality is commonly discussed outside of NLP, it is highly applicable to NLP research but remains largely overlooked.
For example, a U.S.-born English-speaking researcher building a toxicity detection system will likely start with U.S.-centric English statements to annotate for toxicity. 
This can cause the tool to work poorly for other populations (e.g., not detect offensive terms like ``\textit{presstitute}'' in Indian contexts; see Figure~\ref{fig:positionality}). 

Such \textit{design biases} in the creation of datasets and models, i.e., disparities in how well datasets and models work for different populations, stem from factors including latent design choices and the researcher's positionality.
However, they can perpetuate systemic inequalities by imposing one group's standards onto the rest of the world \cite{ghosh2021detecting,gururangan2022whose,blasi-etal-2022-systematic}.
The challenge is that design biases arise from the myriad of design choices made; 
in the context of creating datasets and models, only some of these choices may be documented \cite[e.g., through model cards and data sheets;][]{bender-friedman-2018-data,mitchell2019model,gebru2021datasheets}. Further, many popular deployed models are hidden behind APIs, and thus design biases can only be characterized indirectly (e.g., by observing model behavior).

We introduce \framework, a framework for characterizing design biases and positionality of NLP datasets and models. 
For a given dataset and task, we obtain a wide set of new annotations for a data sample, from a diverse pool of volunteers from various countries and of different backgrounds \cite[recruited through LabintheWild;][]{reinecke2015labinthewild}.
We then quantify design biases by comparing which identities and backgrounds have higher agreement with the original dataset labels or model predictions.
\interfootnotelinepenalty=10000
\framework offers three advantages over other approaches (e.g., paid crowdsourcing or laboratory studies).
First, the demographic diversity of participants on LabintheWild is better than on other crowdsourcing platforms \cite{reinecke2015labinthewild} and in traditional laboratory studies.
Second, the compensation and incentives in our approach rely on a participant's motivation to learn about themselves instead of monetary compensation. This has been shown to result in higher data quality compared to using paid crowdsourcing platforms~\cite{august2019pay}, as well as in opportunities for participant learning~\cite{oliveira2017citizen}. This allows our framework to \textit{continuously collect} new annotations and reflect more up-to-date measurements of design biases for free over long periods of time, compared to one-time paid studies such as in previous works~\cite{sap-etal-2022-annotators,davani2022dealing}.\footnote{To view the most up-to-date results, visit the project page (\redact{\texttt{\href{http://nlpositionality.cs.washington.edu/}{nlpositionality.cs.washington.edu}}}) or Github repository (\redact{\texttt{\href{https://github.com/liang-jenny/nlpositionality}{github.com/liang-jenny/nlpositionality}}}).}
Finally, our approach is dataset- and model-agnostic and can be applied post-hoc to any dataset or model using only instances and their labels or predictions.

We apply \framework to two case studies of NLP tasks---social acceptability and hate speech detection---which are known to exhibit design biases \cite{talat-etal-2022-machine,sap-etal-2022-annotators,ghosh2021detecting}. 
We examine datasets and supervised models related to these tasks as well as general-purpose large language models (i.e., GPT-4).
As of May 25 2023, a total of $16,299$ annotations were collected from $1,096$ annotators from $87$ countries, with an average of $38$ annotations per day.
We discover that the datasets and models we investigate are most aligned with White and educated young people from English-speaking countries, which are a subset of ``WEIRD'' \cite[Western, Educated, Industrialized, Rich, Democratic;][]{henrich2010weirdest} populations. 
We also see that datasets exhibit close alignment with their original annotators, emphasizing the importance of gathering data and annotations from diverse groups.

Our paper highlights the importance of considering design biases in NLP. Our findings showcase the usefulness of our framework in quantifying dataset and model positionality.
In a discussion of the implications of our results, we consider how positionality may manifest in other NLP tasks. 

%% file: sections/02_design_bias.tex
\section{Dataset \& Model Positionality: Definitions and Background}
A person's positionality is the perspectives they hold as a result of their demographics, identity, and life experiences~\cite{holmes2020researcher,savin2013qualititative}. 
For researchers, positionality ``reflects the position that [they have] chosen to adopt within a given research study''~\cite{savin2013qualititative}. It influences the research process and its outcomes and results~\cite{rowe2014positionality}. 
Some aspects of positionality, such as gender, race, skin color, and nationality, are culturally ascribed and part of one's identity. Others, such as political views and life history, are more subjective~\cite{holmes2020researcher,foote2011pathways}. 

\paragraph{Dataset and Model Positionality} 
While positionality is often attributed to a person, in this work, we focus on \textit{dataset and model positionality}. \citet{cambo2022model} introduced model positionality, defining it as ``\textit{the social and cultural position of a model with regard to the stakeholders with which it interfaces}.'' We extend this definition to add that datasets also encode positionality, in a similar way as models. This results in perspectives embedded within language technologies, making them less inclusive towards certain populations. 

\begin{figure}[t]
    \centering
    \includegraphics[width=0.90\columnwidth,page=1]{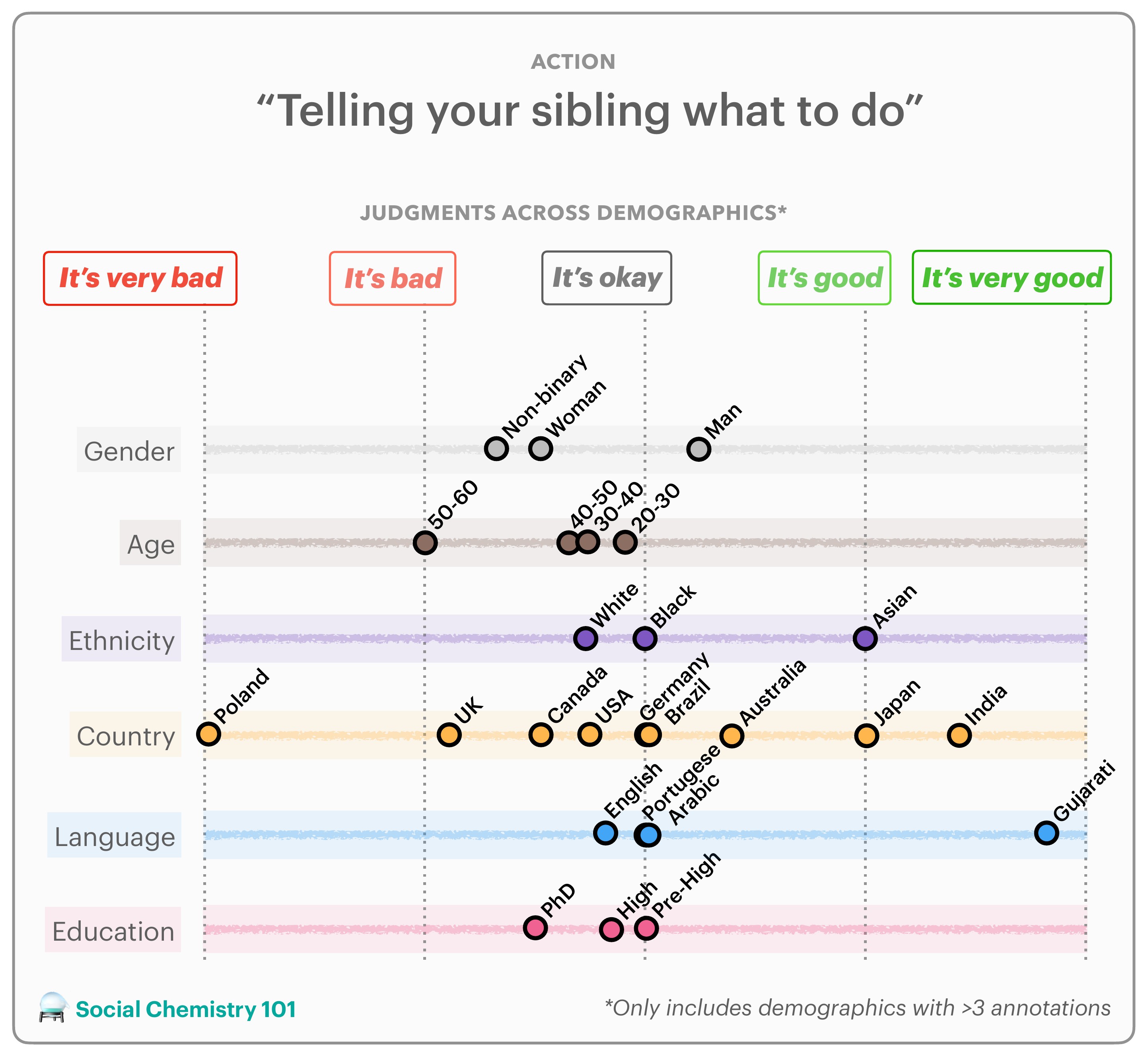}
    \caption{\textbf{Example Annotation.} An example instance from the Social Chemistry dataset that was sent to LabintheWild along with the mean of the received annotation scores across various demographics.
    }
    \label{fig:annotation-fig}
\end{figure}

\paragraph{Design Biases}
In NLP, design biases occur when a researcher or practitioner makes design choices---often based on their positionality---that cause models and datasets to systematically work better for some populations over others. 
Curating datasets involves design choices such as what source to use, what language to use, what perspectives to include or exclude, 
or who to get annotations from. For example, a researcher's native language may influence them to create datasets in that language due to their familiarity with the domain (as in the example in Figure \ref{fig:positionality}).
When training models, these choices include the type of training data, data pre-processing techniques, or the objective function~\cite{hall2022systematic}. For example, a researcher's institutional affiliation may influence the training datasets they select (e.g., choosing a dataset made by a coworker).
Since the latent choices that result in design biases are fundamental to research itself, some researchers have argued that it is impossible to completely de-bias datasets and models~\cite{waseem2021disembodied}.

Current discussions around bias in NLP often focus on ones that originate from social biases embedded within the data. In comparison, design biases originate from the developer who makes assumptions. Based on \citet{friedman1996bias}'s framework on bias, social biases are pre-existing biases in society, whereas design biases are emergent biases that originate from the computing system itself.
`Gender bias' in computing systems means that
the system does not perform well for some genders; ``man is to doctor as woman is to nurse''~\cite{bolukbasi2016man} is a social bias, while captioning systems that fail to understand women's voices~\cite{tatman2017gender} is a design bias.

One prominent example of design bias in NLP is the overt emphasis on English~\cite{joshi-etal-2020-state,blasi-etal-2022-systematic}.
Others include the use of block lists in dataset creation or toxicity classifiers as a filter, which can marginalize minority voices~\cite{dodge-etal-2021-documenting,xu2021detoxifying}. In this work, we extend the discussion of design biases from prior work into NLP, discuss it in relation to researcher positionality, and show its effects on datasets and models.

%% file: sections/03_setup.tex
\begin{figure*}[t]
    \centering
    \includegraphics[width=0.93\textwidth,page=1]{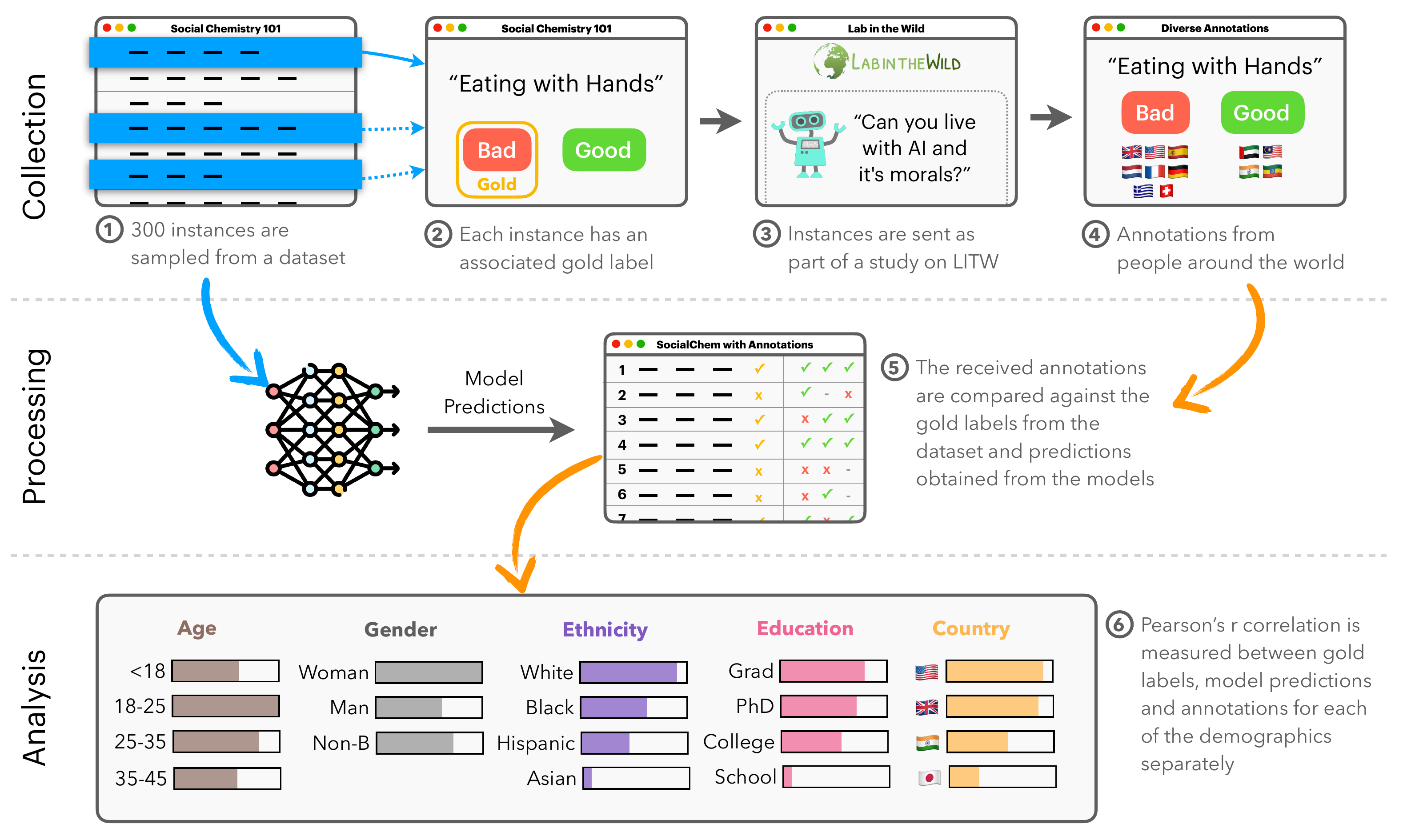}
    \caption{\textbf{Overview of the \framework Framework.} \underline{\smash{Collection (steps 1-4)}}: A subset of datasets' instances are re-annotated via diverse volunteers recruited on LabintheWild. \underline{\smash{Processing (step 5)}}: We compare the labels collected from LabintheWild with the dataset's original labels and models' predictions. \underline{\smash{Analysis (step 6)}}: We compute the Pearson's $r$ correlation between the LabintheWild annotations by demographic for the dataset's original labels and the models' predictions. We apply the Bonferroni correction to account for multiple hypothesis testing.}
    \label{fig:framework-fig}
\end{figure*}

\section{\framework: Quantifying Dataset and Model Positionality}
Our \framework framework follows a two-step process for characterizing the design biases and positionality of datasets and models. 
First, a subset of data for a task is re-annotated by annotators from around the world to obtain globally representative data in order to quantify positionality (\S\ref{ssec:collecting-anns}). We specifically rely on re-annotation to capture self-reported demographic data of annotators with each label.
Then, the positionality of the dataset or model is computed by comparing the responses of the dataset or model with different demographic groups for identical instances (\S\ref{ssec:computing-pos}). While relying on demographics as a proxy for positionality is limited (see discussion in \S\ref{sec:limitations}), we use demographic information for an initial exploration in uncovering design biases in datasets and models.

\subsection{Collecting Diverse Annotations}
\label{ssec:collecting-anns}
Cost-effectively collecting annotations from a diverse crowd at scale is challenging. Popular crowdsourcing platforms like Amazon Mechanical Turk (MTurk) are not culturally diverse, as a majority of workers are from the United States and India~\cite{difallah2018demographics,ipeirotis2010demographics}. Further, MTurk does not easily support continuous and longitudinal data collection. To address these challenges, we use LabintheWild~\cite{reinecke2015labinthewild}, which hosts web-based online experiments. Compared to traditional laboratory settings, it has more diverse participants and collects equally high-quality data for free~\cite{august2019pay,oliveira2017citizen}; instead of monetary compensation, participants typically partake in LabintheWild experiments because they learn something about themselves. %
Thus, we motivate people to participate in our IRB-approved study (\S\ref{sec:ethics}) by enabling them to learn how their responses on a given task (e.g., judging hate speech) compare to a judgment by AI systems as well as by others who are demographically similar to them (see Appendix~\ref{appendix:study-design}). 

For a given task, we choose a dataset to be annotated. To select instances for re-annotation, we filter the dataset based on relevant information that could indicate subjectivity (such as \textit{controversiality} label for the social acceptability dataset), and then sample $300$ diverse instances by stratified sampling across different dataset metadata, (such as the \emph{targeted groups of toxic speech} label for the hate speech dataset) (see Appendix~\ref{appendix:sampling}).
These instances are then hosted as an experiment on LabintheWild to be annotated by a diverse crowd, where participants report their demographics. To ensure consistency in the re-annotated data, the instructions and annotation setups are similar to the original tasks'. Figure~\ref{fig:annotation-fig} is an example from the Social Chemistry dataset and its annotations.

\subsection{Quantifying Positionality}
\label{ssec:computing-pos}
We use correlation as a quantitative construct for positionality. First, we group the annotations by specific demographics. When datasets contain multiple annotations from the same demographic for the same instance, we take the mean of the labels from annotators of that demographic to obtain an aggregated score (see Table~\ref{tab:demographic-differences}).  
Next, for each demographic, we compute Pearson's $r$ using the demographic's aggregated score for each instance and correlated it to the dataset label or model prediction\footnote{We use models' output probability scores for supervised models and categorical labels for GPT-4.}.
We then apply the Bonferroni correction to account for multiple hypotheses testing~\cite{wickens2004design}. We rank the correlations to reveal which demographic groups best align with the positionality of datasets and models.
Finally, we report the total number of annotators and inter-annotator agreements for each demographic using Krippendorff's $\alpha$~\cite{krippendorff2004reliability}.

%% file: sections/04_case_studies.tex
\input{tables/demographic-differences.tex}

\section{Case Studies}
\input{tables/correlation-auto-new.tex}
We present case studies of applying \framework to two different socially-aligned tasks: social acceptability and hate speech detection.

\subsection{Social Acceptability}
Social acceptability is the task of evaluating how acceptable certain actions are in society. Related to making moral judgments, it requires models to have commonsense rules about social norms and acceptable behavior~\cite{forbes-etal-2020-social}.
Social acceptability is likely to vary by background, as morality~\cite{atari2022morality}, values~\cite{haerpfer2012world}, and moral decision-making~\cite{awad2018moral,awad2020universals} vary by culture. 
In NLP, social acceptability models' behavior have been shown to systematically display certain values~\cite{arora2022probing} and biases~\cite{hammerl2022speaking}, some of which are Western-centric~\cite{jiang2021delphi,talat-etal-2022-machine}.

\paragraph{Setup}
$612$ participants annotated the social acceptability of $300$ examples from the Social Chemistry dataset~\cite{forbes-etal-2020-social}. We collected $11,294$ annotations for this task and compared participants' responses to the original Social Chemistry dataset, the Delphi model~\cite{jiang2021delphi}, and GPT-4 zero-shot~\cite{openai2023gpt}. In order to evaluate GPT-4, we combine Delphi's GPT-3 zero-shot evaluation setup from~\citet{jiang2021delphi} with the template used by~\citet{rae2021scaling}:
\begin{examplebox}
\begin{footnotesize}
\begin{verbatim}
Text:     <Social Chemistry action>
Question: How socially acceptable is the above
          situation: good, okay, or bad?
Answer:
\end{verbatim}
\end{footnotesize}
\end{examplebox}

\paragraph{Results}
We find that the instances we selected for the social acceptability task can vary by the annotator's demographics (see Table~\ref{tab:demographic-differences}). For example, men and non-binary people are more likely than women to say it's okay to tell a woman what to do.

We also report the Pearson's $r$ results in Table~\ref{tab:stat_model}, with corresponding $p$-values after applying the Bonferroni correction in the Appendix~\ref{sec:p-values}. We use $^*$ to represent statistically significant Pearson's $r$ coefficients ($p < 2.04e-05$). Social Chemistry is most aligned with people who grow up ($r=0.76^*$) and live in ($r=0.76^*$) English-speaking countries, who have a college education ($r=0.74^*$), are White ($r=0.73^*$), and are 20-30 years old ($r=0.74^*$), indicating a preference to younger WEIRD populations. 

Delphi also exhibits a similar pattern, but to a lesser degree. While it strongly aligns with people who grow up ($r=0.61^*$) and live in ($r=0.65^*)$ English-speaking countries, who have a college education ($r=0.66^*$), are White ($r=0.61^*$), and are 20-30 years old ($r=0.66^8$); it also correlates more with other populations, such as people who grow up ($r=0.72^*$) in Baltic countries compared to English-speaking countries.

We also observe a similar pattern with GPT-4. It has the highest Pearson's $r$ value for people who grow up ($r=0.74^*$) and live in ($r=0.73^*$) English-speaking countries, are college-educated ($r=0.69^*$), are White ($r=0.70^*$) and are between 20-30 years old ($r=0.70^*$). However, it does have strong correlations with other populations, such as people who grow up in Confucian countries ($r=0.74^*$) and have a graduate degree ($r=0.69^*$).

Native English speakers strongly correlate with both datasets and models. 
Non-binary people align less to both Social Chemistry, Delphi, and GPT-4 compared to men and women.
People with professional school education and Black, Latinx, and Native American populations consistently rank least in correlation for education level and ethnicity.

\subsection{Hate Speech Detection}
Hate speech detection is the task of determining whether an instance is ``\textit{abusive speech targeting specific group characteristics,
such as ethnic origin, religion, gender, or sexual
orientation}''~\cite{warner2012detecting}. 
Rating whether text is hate speech is likely to vary by background. 
Also, annotators rate the toxicity of tweets with African American English (AAE) differently by background, such as political affiliation~\cite{sap-etal-2022-annotators}. 
Hate speech classifiers also produce errors for instances outside Western contexts~\cite{ghosh2021detecting}.

\paragraph{Setup}
For the hate speech detection task, $484$ LabintheWild participants annotated $300$ examples from the Dynahate dataset~\cite{vidgen-etal-2021-learning} and rated whether an instance was hate speech or not. We collected $5,005$ annotations and compared participants' responses to the original Dynahate dataset as well as Perspective API\footnote{\texttt{\href{https://perspectiveapi.com/}{perspectiveapi.com}}}, Rewire API\footnote{\texttt{\href{https://rewire.online/}{rewire.online}}}, HateRoBERTa~\cite{hartvigsen-etal-2022-toxigen}, and GPT-4 zero-shot. To evaluate GPT-4, we use the prompt used by~\citet{rae2021scaling}:

\begin{examplebox}
\begin{footnotesize}
\begin{verbatim}
Text:     <Dynahate instance>
Question: Does the above text contain rude, 
          hateful, aggressive, disrespectful,
          or unreasonable language?
Answer:
\end{verbatim}
\end{footnotesize}
\end{examplebox}

\paragraph{Results}
We find that the instances we selected for this task can vary by the annotator's demographics (see Table~\ref{tab:demographic-differences}). For instance, annotators who were racial minorities rated the instance ``\emph{Justice for the yellow man lol}'' as hate speech, while White annotators voted being not sure.

We also report the results in Table~\ref{tab:stat_model}. We report the $p$-values after applying the Bonferroni correction in the Appendix~\ref{sec:p-values}. Dynahate is highly correlated with people who grow up in English-speaking countries ($r=0.70^*$), who have a college education ($r=0.66^*$), are White ($r=0.69^*$), and are 20-30 years old ($r=0.65^*$). However, it also has high alignment with other populations, such as people who live in West South Asia ($r=0.77$).

Perspective API also tends to align with WEIRD populations, though to a lesser degree than Dynahate. Perspective API exhibits some alignment with people who grow up and live in English-speaking ($r=0.33^*$, $r=0.34^*$ respectively), have a college education ($r=0.34^*$), are White ($r=0.29^*$), and are 20-30 years old ($r=0.34^*$). It also exhibits higher alignment with other populations, such as people who live in Confucian countries ($r=0.36$) compared to English-speaking countries. Unexpectedly, White people rank lowest in Pearson's $r$ score within the ethnicity category.

Rewire API similarly shows this bias. It has a moderate correlation with people who grow up and live in English-speaking countries ($r=0.58^*$, $r=0.60^*$ respectively), have a college education ($r=0.56^*$), are White ($r=0.56^*$), and are 20-30 years old ($r=0.56^*$).

A Western bias is also shown in HateRoBERTa. HateRoBERTa shows some alignment with people who grow up ($r=0.37^*$) and live in ($r=0.38^*$) English-speaking countries, have a college education ($r=0.38^*$), are White ($r=0.32^*$), and are between 20-30 years of age ($r=0.38^*$). 

We also observe similar behavior with GPT-4. The demographics with some of the higher Pearson's $r$ values in its category are people who grow up ($r=0.41^*$) and live in ($r=0.42^*$) English-speaking countries, are college-educated ($r=0.39^*$), are White ($r=0.38^*$), and are 20-30 years old ($r=0.42^*$). It shows stronger alignment to Asian-Americans ($r=0.39^*$) compared to White people, as well as people who live in Baltic countries ($r=0.75$) and people who grow up in Confucian countries ($r=0.52^*$) compared to people from English-speaking countries.

As in the previous task, labels from native English speakers are strongly correlated with datasets and models. Non-binary people align less with Dynahate, Perspective API, Rewire, HateRoBERTa, and GPT-4 compared to other genders.
Also, people who are professional school-educated or are Black, Latinx, and Native American rank least in alignment for education and ethnicity respectively.

%% file: tables/demographic-differences.tex
\newcommand{\spsize}{0.05cm}
\newcommand{\centered}[1]{\begin{tabular}{l} #1 \end{tabular}}
\newcommand{\q}{\phantom{*}}

\definecolor{fade}{HTML}{111111}

\newcommand*{\mline}[1]{%
\begingroup
    \renewcommand*{\arraystretch}{1.1}%
   \begin{tabular}[c]{@{}>{\raggedright\arraybackslash}p{5.75cm}@{}}#1\end{tabular}%
  \endgroup
}

\begin{table*}[]
\footnotesize
    \centering
    \begin{tabular}{p{0.15cm}p{5.75cm}p{3.5cm}p{2cm}p{1.5cm}p{1cm}}
    \toprule
    & \textbf{Instance} & \textbf{Demographic} & \textbf{Label} & \mline{\textbf{$\mu$ Score}}  & \textbf{$\sigma^{2}$} \\
    \midrule
       \textbf{\multirow{4}{*}[0ex]{\rotatebox[origin=c]{90}{Social Acceptability~}}} & \textcolor{fade}{\fontfamily{lmtt}\selectfont Telling a woman what to do.} & \mline{Man \newline Non-binary \newline Woman} & \mline{It's okay \newline It's okay \newline It's bad} & \mline{\acc{-0.38} \newline \acc{-0.33} \newline \acc{-1.33}} & \mline{\varz{0.55} \newline \varz{4.33} \newline \varz{0.27}} \\

        \cdashline{2-6}[1.5pt/3pt]\noalign{\vskip \spsize}
        
        & \mline{\textcolor{fade}{\fontfamily{lmtt}\selectfont Questioning your sexual orientation.}} & \mline{10-20 years old \newline 20-30 years old \newline 30-40, 40-50 years old} & \mline{It's good \newline It's okay \newline It's okay} & \mline{ \acc{0.88}\newline \acc{0.14} \newline \acc{0.00}} & \mline{\varz{0.70} \newline \varz{1.48} \newline \varz{0.00}} \\

       \cdashline{2-6}[1.5pt/3pt]\noalign{\vskip \spsize}
       & \mline{\textcolor{fade}{\fontfamily{lmtt}\selectfont Going to church to practice your religion.}} & \mline{Christian \newline Buddhist \newline None} & \mline{It's good \newline It's okay \newline It's okay} & \mline{\acc{1.33} \newline \acc{0.00} \newline \acc{0.05}} & \mline{\varz{1.33} \newline \varz{0.00} \newline \varz{0.50}} \\

        \midrule
        \textbf{\multirow{3}{*}[-2.3ex]{\rotatebox[origin=c]{90}{Hate Speech~}}} 
        & \textcolor{fade}{\fontfamily{lmtt}\selectfont Justice for yellow man lol} & \mline{Asian, Black, Latino/Latina \newline White} & \mline{Hate speech \newline Not sure} & \mline{\tox{1.00} \newline \tox{0.33}} & \mline{\varz{0.00} \newline \varz{1.33}}\\

        \cdashline{2-6}[1.5pt/3pt]\noalign{\vskip \spsize}
        & \mline{\textcolor{fade}{\fontfamily{lmtt}\selectfont Iran's Supreme Court claims gender equality is `Zionist plot' aiming to corrupt role of women in society. I fucking second that!}} & \mline{Christian \newline Jewish \newline Muslim} & \mline{Not sure \newline Hate speech \newline Not hate speech} & \mline{\tox{0.50} \newline \tox{1.00} \newline \tox{-1.00}} & \mline{\varz{0.92} \newline \varz{0.00} \newline \varz{0.00}}\\

       \cdashline{2-6}[1.5pt/3pt]\noalign{\vskip \spsize}
       
       & \mline{\textcolor{fade}{\fontfamily{lmtt}\selectfont Literally anyone can be a shitty person, LGBT people especially!}} & \mline{College, Graduate school \newline Pre-high school} & \mline{Hate speech \newline Not hate speech} & \mline{\tox{1.00} \newline \tox{-1.00}} & \mline{\varz{0.00} \newline \varz{0.00}} \\
       
    \bottomrule
    \end{tabular}
    \caption{\textbf{Disagreement Examples.} Examples where there was disagreement within and between demographics on what the label should be from LabintheWild participants. Disagreement is shown for each label through the variance ($\sigma^2$) and the mean ($\mu$) of obtained annotations for a particular demographic.}
    \label{tab:demographic-differences}
\end{table*}

%% file: tables/correlation-auto-new.tex
\input{tables/aa_variables.tex}
\begin{table*}[]
\scriptsize
    \centering
    \begin{tabular}{p{3.8cm}"p{0.4cm}p{0.52cm}|>{\columncolor{datasetcolor}}P{0.56cm}P{0.56cm}P{0.56cm}"p{0.4cm}p{0.52cm}|>{\columncolor{datasetcolor}}P{0.56cm}P{0.56cm}P{0.56cm}P{0.56cm}P{0.56cm}"}
    \toprule
    \multicolumn{13}{l}{
    {\fontfamily{lmtt}\selectfont
     DATASETS:} {\scriptsize \socialchemlogo SocialChemistry \quad \dynahatelogo DynaHate} \hfill {\fontfamily{lmtt}\selectfont MODELS:} {\scriptsize \gptlogo GPT-4 \quad \delphilogo Delphi \quad \perspectivelogo PerspectiveAPI \quad \rewirelogo RewireAPI \quad \hatebertlogo HateRoberta}} \\
    \toprule
      \textbf{Demographic} &  \multicolumn{12}{c"}{\textbf{Pearson's $r$}}\\
    \midrule
    & \multicolumn{5}{c"}{\ul{\textbf{Social Acceptability}}} & \multicolumn{7}{c"}{\ul{\textbf{Toxicity \& Hate Speech}}}\\[\tabsp]
    & \acceptabilitynumlogo & \acceptabilityirrlogo & \socialchemlogo & \delphilogo & \gptacclogo & \toxicitynumlogo & \toxicityirrlogo & \dynahatelogo & \perspectivelogo & \rewirelogo & \hatebertlogo & \gpttoxlogo\\
    \cmidrule{2-13}
\sethlcolor{countrycolor} 
 \textbf{\codebox{} Country (Lived Longest)} \dotfilla & \multicolumn{12}{c"}{\dotfilla\dotfilla\dotfilla\dotfilla\dotfilla\dotfilla\dotfilla\dotfilla\dotfilla\dotfilla\dotfilla\dotfilla\dotfilla\dotfilla\dotfilla\dotfilla\dotfilla\dotfilla\dotfilla\dotfilla\dotfilla\dotfilla\dotfilla\dotfilla\dotfilla\dotfilla\dotfilla\dotfilla\dotfilla\dotfilla\dotfilla\dotfilla\dotfilla\dotfilla\dotfilla\dotfilla\dotfilla\dotfilla\dotfilla\dotfilla\dotfilla\dotfilla\dotfilla\dotfilla\dotfilla} \\ 
\quad African Islamic & \metanum{316} & \metanum{0.20\q} & 0.54* & 0.49\q & \minnum{0.47\q} & \metanum{234} & \metanum{0.22\q} & 0.39\q & 0.29\q & 0.39\q & 0.27\q & 0.25\q \\
\quad Baltic & \metanum{140} & \metanum{0.41\q} & 0.73* & \maxnum{0.72*} & 0.71* & \metanum{54} & \metanum{0.50\q} & 0.38\q & \minnum{-0.08} & \minnum{0.20\q} & \minnum{0.05\q} & 0.46\q \\
\quad Catholic Europe & \metanum{452} & \metanum{0.28\q} & 0.64* & 0.59* & 0.68* & \metanum{183} & \metanum{0.41\q} & \minnum{0.32\q} & 0.12\q & 0.32\q & 0.21\q & 0.21\q \\
\quad Confucian & \metanum{528} & \metanum{0.42\q} & 0.75* & 0.58* & \maxnum{0.74*} & \metanum{154} & \metanum{0.24\q} & 0.47\q & 0.28\q & 0.51* & 0.12\q & \maxnum{0.52*} \\
\quad English-Speaking & \metanum{8289} & \metanum{0.51\q} & \maxnum{0.76*} & 0.61* & \maxnum{0.74*} & \metanum{4025} & \metanum{0.40\q} & \maxnum{0.70*} & \maxnum{0.33*} & \maxnum{0.58*} & \maxnum{0.37*} & 0.41* \\
\quad Latin American & \metanum{281} & \metanum{0.33\q} & \minnum{0.45\q} & \minnum{0.41\q} & \minnum{0.47\q} & \metanum{65} & \metanum{0.20\q} & 0.39\q & 0.10\q & 0.28\q & 0.09\q & 0.17\q \\
\quad Orthodox Europe & \metanum{426} & \metanum{0.39\q} & 0.56* & 0.58* & 0.67* & \metanum{139} & \metanum{0.32\q} & 0.36\q & 0.18\q & 0.47\q & 0.15\q & \minnum{0.13\q} \\
\quad Protestant Europe & \metanum{706} & \metanum{0.48} & 0.65* & 0.57* & 0.67* & \metanum{387} & \metanum{0.37} & 0.40* & 0.32\q & 0.23\q & 0.29\q & 0.34\q \\
\quad West South Asia & \metanum{413} & \metanum{0.40\q} & 0.63* & 0.60* & 0.59* & \metanum{116} & \metanum{0.21} & 0.34\q & 0.20\q & 0.33\q & 0.30\q & 0.21\q \\
\sethlcolor{educationcolor} 
 \textbf{\codebox{} Education Level} \dotfilla & \multicolumn{12}{c"}{\dotfilla\dotfilla\dotfilla\dotfilla\dotfilla\dotfilla\dotfilla\dotfilla\dotfilla\dotfilla\dotfilla\dotfilla\dotfilla\dotfilla\dotfilla\dotfilla\dotfilla\dotfilla\dotfilla\dotfilla\dotfilla\dotfilla\dotfilla\dotfilla\dotfilla\dotfilla\dotfilla\dotfilla\dotfilla\dotfilla\dotfilla\dotfilla\dotfilla\dotfilla\dotfilla\dotfilla\dotfilla\dotfilla\dotfilla\dotfilla\dotfilla\dotfilla\dotfilla\dotfilla\dotfilla} \\ 
\quad College & \metanum{4489} & \metanum{0.48} & \maxnum{0.74*} & \maxnum{0.66*} & \maxnum{0.69*} & \metanum{2383} & \metanum{0.39} & \maxnum{0.66*} & \maxnum{0.34*} & \maxnum{0.56*} & \maxnum{0.38*} & \maxnum{0.39*} \\
\quad Graduate School & \metanum{1116} & \metanum{0.53} & 0.72* & 0.54* & \maxnum{0.69*} & \metanum{604} & \metanum{0.36} & 0.59* & 0.28* & 0.51* & 0.25\q & 0.38* \\
\quad High School & \metanum{2183} & \metanum{0.49} & 0.67* & 0.54* & 0.64* & \metanum{908} & \metanum{0.41} & 0.60* & 0.25\q & 0.49* & 0.30* & 0.37* \\
\quad PhD & \metanum{709} & \metanum{0.46} & 0.65* & 0.55* & 0.61* & \metanum{359} & \metanum{0.45} & 0.48* & 0.19\q & 0.43* & 0.26\q & 0.31\q \\
\quad Pre-High School & \metanum{406} & \metanum{0.40\q} & 0.56* & \minnum{0.46*} & 0.59* & \metanum{116} & \metanum{0.26} & \minnum{0.37\q} & 0.24\q & 0.45* & 0.25\q & 0.38\q \\
\quad Professional School & \metanum{460} & \metanum{0.40\q} & \minnum{0.53*} & \minnum{0.46*} & \minnum{0.49*} & \metanum{195} & \metanum{0.09} & 0.61* & \minnum{0.10\q} & \minnum{0.35\q} & \minnum{0.09\q} & \minnum{0.19\q} \\

\sethlcolor{ethnicitycolor} 
 \textbf{\codebox{} Ethnicity} \dotfilla & \multicolumn{12}{c"}{\dotfilla\dotfilla\dotfilla\dotfilla\dotfilla\dotfilla\dotfilla\dotfilla\dotfilla\dotfilla\dotfilla\dotfilla\dotfilla\dotfilla\dotfilla\dotfilla\dotfilla\dotfilla\dotfilla\dotfilla\dotfilla\dotfilla\dotfilla\dotfilla\dotfilla\dotfilla\dotfilla\dotfilla\dotfilla\dotfilla\dotfilla\dotfilla\dotfilla\dotfilla\dotfilla\dotfilla\dotfilla\dotfilla\dotfilla\dotfilla\dotfilla\dotfilla\dotfilla\dotfilla\dotfilla} \\ 
\quad Asian, Asian American & \metanum{1160} & \metanum{0.55} & 0.66* & 0.55* & 0.63* & \metanum{644} & \metanum{0.45} & 0.57* & 0.35* & 0.47* & 0.33* & 0.39* \\
\quad Black, African American & \metanum{465} & \metanum{0.52} & 0.61* & \minnum{0.50*} & 0.57* & \metanum{287} & \metanum{0.34\q} & 0.56* & 0.32\q & 0.36* & \minnum{0.31\q} & 0.37* \\
\quad Latino / Latina, Hispanic & \metanum{314} & \metanum{0.57} & 0.62* & 0.52* & \minnum{0.54*} & \metanum{239} & \metanum{0.36} & 0.43* & 0.39* & 0.46* & \minnum{0.31\q} & \minnum{0.31\q} \\
\quad Native American, Alaskan Native & \metanum{103} & \metanum{0.64} & \minnum{0.59*} & 0.52* & 0.64* & \metanum{65} & \metanum{---} & \minnum{0.23\q} & 0.31\q & \minnum{0.31\q} & 0.32\q & 0.33\q \\
\quad Pacific Islander, Native Australian & \metanum{38} & \metanum{0} & 0.65* & \maxnum{0.63\q} & 0.62\q & \metanum{27} & \metanum{---} & 0.36\q & \maxnum{0.65\q} & 0.54\q & \maxnum{0.64\q} & \maxnum{0.57\q} \\
\quad White & \metanum{3102} & \metanum{0.55} & \maxnum{0.73*} & 0.61* & \maxnum{0.70*} & \metanum{1831} & \metanum{0.44} & \maxnum{0.69*} & \minnum{0.29*} & \maxnum{0.56*} & 0.32* & 0.38* \\

\sethlcolor{gendercolor} 
 \textbf{\codebox{} Gender} \dotfilla & \multicolumn{12}{c"}{\dotfilla\dotfilla\dotfilla\dotfilla\dotfilla\dotfilla\dotfilla\dotfilla\dotfilla\dotfilla\dotfilla\dotfilla\dotfilla\dotfilla\dotfilla\dotfilla\dotfilla\dotfilla\dotfilla\dotfilla\dotfilla\dotfilla\dotfilla\dotfilla\dotfilla\dotfilla\dotfilla\dotfilla\dotfilla\dotfilla\dotfilla\dotfilla\dotfilla\dotfilla\dotfilla\dotfilla\dotfilla\dotfilla\dotfilla\dotfilla\dotfilla\dotfilla\dotfilla\dotfilla\dotfilla} \\ 
\quad Man & \metanum{4082} & \metanum{0.45} & 0.73* & \maxnum{0.63*} & 0.69* & \metanum{1798} & \metanum{0.37} & \maxnum{0.65*} & \maxnum{0.34*} & \maxnum{0.56*} & 0.34* & 0.36* \\
\quad Non-Binary & \metanum{858} & \metanum{0.41} & \minnum{0.60*} & \minnum{0.51*} & \minnum{0.55*} & \metanum{329} & \metanum{0.48} & \minnum{0.57*} & \minnum{0.21\q} & \minnum{0.37*} & \minnum{0.27\q} & \minnum{0.31*} \\
\quad Woman & \metanum{4368} & \metanum{0.55} & \maxnum{0.74*} & 0.60* & \maxnum{0.73*} & \metanum{2357} & \metanum{0.39} & 0.63* & \maxnum{0.34*} & 0.53* & \maxnum{0.38*} & \maxnum{0.37*} \\

\sethlcolor{languagecolor} 
 \textbf{\codebox{} Native Language} \dotfilla & \multicolumn{12}{c"}{\dotfilla\dotfilla\dotfilla\dotfilla\dotfilla\dotfilla\dotfilla\dotfilla\dotfilla\dotfilla\dotfilla\dotfilla\dotfilla\dotfilla\dotfilla\dotfilla\dotfilla\dotfilla\dotfilla\dotfilla\dotfilla\dotfilla\dotfilla\dotfilla\dotfilla\dotfilla\dotfilla\dotfilla\dotfilla\dotfilla\dotfilla\dotfilla\dotfilla\dotfilla\dotfilla\dotfilla\dotfilla\dotfilla\dotfilla\dotfilla\dotfilla\dotfilla\dotfilla\dotfilla\dotfilla} \\ 
\quad English & \metanum{7338} & \metanum{0.51} & \maxnum{0.76*} & \maxnum{0.64*} & \maxnum{0.71*} & \metanum{3622} & \metanum{0.40\q} & \maxnum{0.70*} & \maxnum{0.33*} & \maxnum{0.60*} & \maxnum{0.39*} & \maxnum{0.42*} \\
\quad Not English & \metanum{2157} & \metanum{0.40\q} & \minnum{0.62*} & \minnum{0.54*} & \minnum{0.64*} & \metanum{1020} & \metanum{0.27} & \minnum{0.46*} & \minnum{0.32*} & \minnum{0.39*} & \minnum{0.32*} & \minnum{0.36*} \\

\sethlcolor{agecolor} 
 \textbf{\codebox{} Age} \dotfilla & \multicolumn{12}{c"}{\dotfilla\dotfilla\dotfilla\dotfilla\dotfilla\dotfilla\dotfilla\dotfilla\dotfilla\dotfilla\dotfilla\dotfilla\dotfilla\dotfilla\dotfilla\dotfilla\dotfilla\dotfilla\dotfilla\dotfilla\dotfilla\dotfilla\dotfilla\dotfilla\dotfilla\dotfilla\dotfilla\dotfilla\dotfilla\dotfilla\dotfilla\dotfilla\dotfilla\dotfilla\dotfilla\dotfilla\dotfilla\dotfilla\dotfilla\dotfilla\dotfilla\dotfilla\dotfilla\dotfilla\dotfilla} \\ 
\quad 10-20 yrs old & \metanum{3360} & \metanum{0.50\q} & 0.70* & 0.61* & 0.69* & \metanum{1615} & \metanum{0.39} & 0.61* & 0.32* & 0.55* & 0.36* & 0.36* \\
\quad 20-30 yrs old & \metanum{4066} & \metanum{0.47} & \maxnum{0.74*} & \maxnum{0.66*} & \maxnum{0.70*} & \metanum{2114} & \metanum{0.39} & \maxnum{0.65*} & 0.34* & 0.56* & \maxnum{0.38*} & 0.42* \\
\quad 30-40 yrs old & \metanum{870} & \metanum{0.51} & 0.66* & 0.52* & 0.61* & \metanum{419} & \metanum{0.28} & \minnum{0.48*} & 0.14\q & 0.41* & 0.24\q & 0.29\q \\
\quad 40-50 yrs old & \metanum{655} & \metanum{0.44} & 0.62* & 0.55* & 0.63* & \metanum{256} & \metanum{0.28} & 0.63* & 0.29\q & \maxnum{0.57*} & 0.31\q & 0.37* \\
\quad 50-60 yrs old & \metanum{308} & \metanum{0.49} & 0.69* & 0.53* & 0.60* & \metanum{199} & \metanum{0.39\q} & 0.57* & 0.26\q & 0.41* & 0.20\q & 0.25\q \\
\quad 60-70 yrs old & \metanum{204} & \metanum{0.48} & 0.64* & 0.49* & 0.60* & \metanum{19} & \metanum{---} & 0.57\q & \maxnum{0.42\q} & 0.46\q & 0.05\q & \minnum{-0.18} \\
\quad 70-80 yrs old & \metanum{68} & \metanum{---} & 0.56* & 0.52* & 0.56* & \metanum{24} & \metanum{---} & 0.50\q & 0.35\q & \minnum{0.36\q} & 0.24\q & \maxnum{0.85*} \\
\quad 80+ yrs old & \metanum{24} & \metanum{---} & \minnum{0.52\q} & \minnum{0.48\q} & \minnum{0.48\q} & \metanum{12} & \metanum{---} & 0.63\q & \minnum{0.01\q} & 0.45\q & \minnum{-0.09} & 0.43\q \\

\sethlcolor{countrycolor} 
 \textbf{\codebox{} Country (Residence)} \dotfilla & \multicolumn{12}{c"}{\dotfilla\dotfilla\dotfilla\dotfilla\dotfilla\dotfilla\dotfilla\dotfilla\dotfilla\dotfilla\dotfilla\dotfilla\dotfilla\dotfilla\dotfilla\dotfilla\dotfilla\dotfilla\dotfilla\dotfilla\dotfilla\dotfilla\dotfilla\dotfilla\dotfilla\dotfilla\dotfilla\dotfilla\dotfilla\dotfilla\dotfilla\dotfilla\dotfilla\dotfilla\dotfilla\dotfilla\dotfilla\dotfilla\dotfilla\dotfilla\dotfilla\dotfilla\dotfilla\dotfilla\dotfilla} \\ 
\quad African Islamic & \metanum{164} & \metanum{0.27\q} & 0.49\q & 0.48\q & 0.46\q & \metanum{116} & \metanum{0.21\q} & 0.35\q & 0.23\q & 0.29\q & 0.15\q & \minnum{0.16\q} \\
\quad Baltic & \metanum{53} & \metanum{0.02} & 0.65\q & \maxnum{0.65\q} & \minnum{0.33\q} & \metanum{14} & \metanum{0.00} & 0.42\q & 0.14\q & 0.52\q & 0.35\q & \maxnum{0.75\q} \\
\quad Catholic Europe & \metanum{406} & \metanum{0.33} & 0.53* & \minnum{0.41*} & 0.64* & \metanum{172} & \metanum{0.37} & 0.32\q & \minnum{0.11\q} & 0.38\q & 0.15\q & 0.22\q \\
\quad Confucian & \metanum{268} & \metanum{0.42} & 0.68* & 0.55* & \maxnum{0.77*} & \metanum{83} & \metanum{0.17\q} & 0.41\q & \maxnum{0.36\q} & 0.45\q & 0.33\q & 0.48\q \\
\quad English-Speaking & \metanum{7315} & \metanum{0.50\q} & \maxnum{0.76*} & \maxnum{0.65*} & 0.73* & \metanum{3819} & \metanum{0.40\q} & 0.72* & 0.34* & \maxnum{0.60*} & \maxnum{0.38*} & 0.42* \\
\quad Latin American & \metanum{166} & \metanum{0.43} & 0.54* & 0.56* & 0.59* & \metanum{53} & \metanum{0.15} & 0.30\q & 0.12\q & 0.26\q & \minnum{-0.04} & 0.17\q \\
\quad Orthodox Europe & \metanum{264} & \metanum{0.38} & \minnum{0.47\q} & 0.57* & 0.60* & \metanum{90} & \metanum{0.31\q} & \minnum{0.25\q} & 0.28\q & 0.37\q & 0.29\q & 0.17\q \\
\quad Protestant Europe & \metanum{736} & \metanum{0.46\q} & 0.63* & 0.57* & 0.61* & \metanum{387} & \metanum{0.36\q} & 0.45* & 0.31\q & \minnum{0.23\q} & 0.31\q & 0.31\q \\
\quad West South Asia & \metanum{166} & \metanum{0.44} & 0.61* & 0.57* & 0.53* & \metanum{21} & \metanum{---} & \maxnum{0.77\q} & 0.22\q & 0.57\q & 0.07\q & \minnum{0.16\q} \\

\sethlcolor{religioncolor} 
 \textbf{\codebox{} Religion} \dotfilla & 
 \multicolumn{12}{c"}{\dotfilla\dotfilla\dotfilla\dotfilla\dotfilla\dotfilla\dotfilla\dotfilla\dotfilla\dotfilla\dotfilla\dotfilla\dotfilla\dotfilla\dotfilla\dotfilla\dotfilla\dotfilla\dotfilla\dotfilla\dotfilla\dotfilla\dotfilla\dotfilla\dotfilla\dotfilla\dotfilla\dotfilla\dotfilla\dotfilla\dotfilla\dotfilla\dotfilla\dotfilla\dotfilla\dotfilla\dotfilla\dotfilla\dotfilla\dotfilla\dotfilla\dotfilla\dotfilla\dotfilla\dotfilla} \\ 
\quad Buddhist & \metanum{189} & \metanum{0.33} & 0.64* & 0.58* & \minnum{0.55*} & \metanum{69} & \metanum{0.40\q} & 0.48\q & 0.10\q & 0.25\q & 0.19\q & \maxnum{0.50\q} \\
\quad Christian & \metanum{1969} & \metanum{0.50\q} & \maxnum{0.73*} & \minnum{0.55*} & \maxnum{0.73*} & \metanum{1080} & \metanum{0.29} & 0.56* & \maxnum{0.34*} & \maxnum{0.49*} & \maxnum{0.36*} & 0.34* \\
\quad Hindu & \metanum{201} & \metanum{0.75} & 0.65* & \maxnum{0.60*} & 0.58* & \metanum{109} & \metanum{0.46\q} & 0.63* & \maxnum{0.34\q} & 0.41\q & 0.30\q & 0.38\q \\
\quad Jewish & \metanum{204} & \metanum{0.50\q} & 0.66* & \maxnum{0.60*} & 0.60* & \metanum{144} & \metanum{0.45} & \maxnum{0.64*} & 0.29\q & 0.43* & 0.29\q & 0.33\q \\
\quad Muslim & \metanum{319} & \metanum{0.36\q} & 0.63* & 0.59* & 0.72* & \metanum{89} & \metanum{0.33} & 0.42\q & 0.16\q & 0.29\q & \minnum{0.14\q} & 0.31\q \\
\quad Spritual & \metanum{88} & \metanum{0.48\q} & \minnum{0.61*} & \maxnum{0.60*} & 0.72* & \metanum{13} & \metanum{---} & \minnum{0.35\q} & \minnum{-0.16} & \minnum{0.15\q} & 0.20\q & --- \\

    \bottomrule
    \end{tabular}
    \caption{\textbf{Positionality of NLP datasets and models} quantified using Pearson's $r$ correlation coefficients. \newText{\# denotes the number of annotations associated with a demographic group. $\alpha$ denotes Krippendorff's alpha of a demographic group for a task. * denotes statistical significance ($p < 2.04e-05$ after Bonferroni correction). For each dataset or model, we denote the minimum and maximum Pearson's $r$ value for in demographic category in red (\minnum{X}) and blue (\maxnum{X}) respectively.}}
    \label{tab:stat_model}
\end{table*}

%% file: tables/aa_variables.tex
\newcommand*{\codebox}[1]{{\hyphenchar\font=45\relax\hl{~#1~}}}

\definecolor{agecolor}{RGB}{141,110,99}
\definecolor{gendercolor}{HTML}{bdbdbd}
\definecolor{religioncolor}{RGB}{38,166,154}
\definecolor{languagecolor}{RGB}{66,165,245}
\definecolor{educationcolor}{HTML}{f06292}
\definecolor{ethnicitycolor}{RGB}{127,86,194}
\definecolor{countrycolor}{HTML}{ffb74d}

\newcommand{\aicon}[1]{{\includegraphics[height=1.6\fontcharht\font`\B]{#1}}\xspace}

\newcommand{\acceptabilitynumlogo}{\#}
\newcommand{\acceptabilityirrlogo}{$\alpha$}

\newcommand{\socialchemlogo}{\aicon{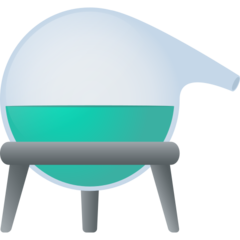}}
\newcommand{\delphilogo}{\aicon{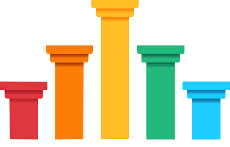}}
\newcommand{\gptlogo}{\aicon{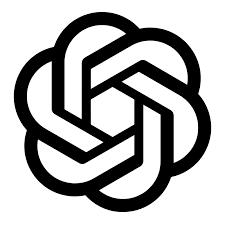}}

\newcommand{\gptacclogo}{\aicon{images/logos/datasets/gpt.png}}
\newcommand{\gpttoxlogo}{\aicon{images/logos/datasets/gpt.png}}

\newcommand{\toxicitynumlogo}{\#}
\newcommand{\toxicityirrlogo}{$\alpha$}

\newcommand{\dynahatelogo}{\aicon{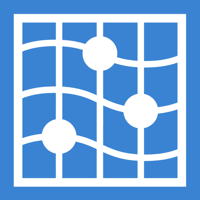}}
\newcommand{\perspectivelogo}{\aicon{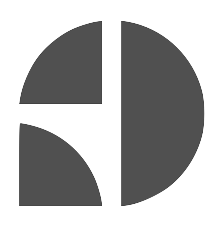}}
\newcommand{\rewirelogo}{\aicon{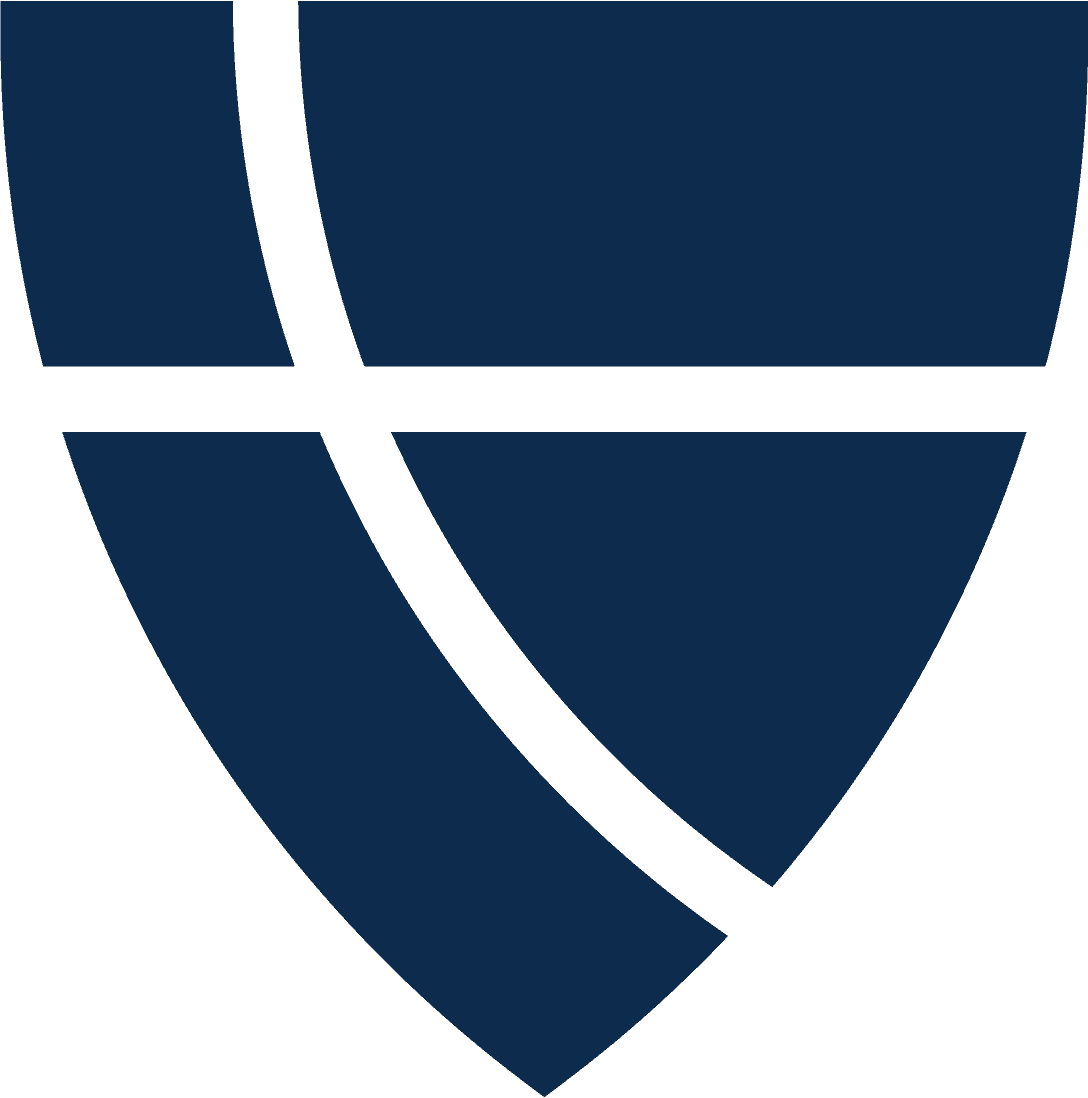}}
\newcommand{\hatebertlogo}{\aicon{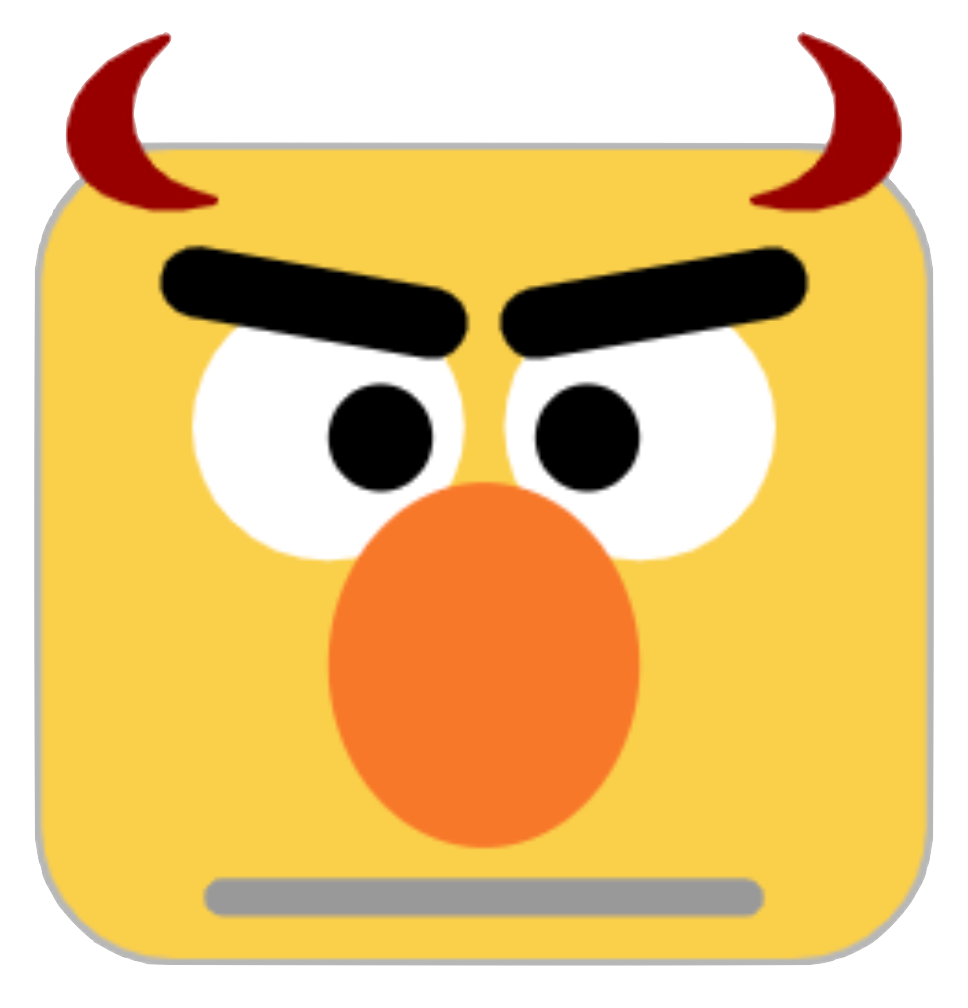}}

\newcommand{\stanfordlogo}{\aicon{images/logos/datasets/stanford.png}}
\newcommand{\politebertlogo}{\aicon{images/logos/datasets/bert.png}}

\newcommand{\dotfilla}{\textcolor{gray!70}{\dotfill}}

\newcommand{\tabsp}[0]{0.15cm}

\newcolumntype{P}[1]{>{\centering\arraybackslash}p{#1}}

\definecolor{stdcolor}{HTML}{757575}
\definecolor{metacolor}{HTML}{757575}

\definecolor{maxcolor}{HTML}{1565c0}
\definecolor{mincolor}{HTML}{e57373}

\definecolor{datasetcolor}{HTML}{eeeeee}

\newcommand{\var}[1]{\textcolor{stdcolor}{\vskip -2.5mm #1}}
\newcommand{\maxnum}[1]{\textbf{\textcolor{maxcolor}{#1}}}
\newcommand{\minnum}[1]{\textbf{\textcolor{mincolor}{#1}}}
\newcommand{\metanum}[1]{\textcolor{metacolor}{#1}}

%% file: sections/05_discussion.tex
\section{Discussion}
\label{sec:discussion}
In this paper, we characterized design biases and the positionality of datasets and models in NLP. We introduced the \framework framework for identifying design biases in NLP datasets and models. 
\framework consists of a two-step process of collecting annotations from diverse annotators for a specific task and then computing the alignment of the annotations to dataset labels and model predictions using Pearson's $r$. We applied \framework to two tasks: social acceptability and hate speech detection, with two datasets and five models in total. In this section, we discuss key takeaways from our experiments and offer recommendations to account for design biases in datasets and models.

\paragraph{There Is Positionality in NLP}
Models and datasets have positionality, as they align better with some populations than others.
This corroborates work from \citet{cambo2022model} on model positionality, which quantifies positionality by inspecting the content of annotated documents, as well as work from \citet{rogers2021changing}, who argues that collecting a corpus of speech inherently encodes a particular world view (e.g., via linguistic structures, topic of conversations, and the speaker's social context).
We extend these works by showing design biases and quantifying dataset and model positionality by computing correlations between LabintheWild annotations, dataset labels, and model predictions.

Our case studies show examples of positionality in NLP. However, most socially-aligned tasks may encode design biases due to differences in language use between demographic groups, for example, commonsense reasoning~\cite{shwartz-2022-good}, question answering~\cite{gor-etal-2021-toward}, and sentiment analysis~\cite{mohamed2022artelingo}. Even tasks that are considered purely linguistic have seen design biases: in parsing and tagging, performance differences exist between texts written by people of different genders~\cite{garimella-etal-2019-womens}, ages~\cite{hovy-sogaard-2015-tagging}, and races~\cite{johannsen-etal-2015-cross,jorgensen-etal-2015-challenges}. This shows how common design biases are in NLP, as language is a social construct~\cite{burr2015social} and technologies are imbued with their creator's values~\cite{friedman1996value}. This raises the question of whether there are any value-neutral language technologies~\cite{birhane2022values,winner2017artifacts}.

\paragraph{Datasets and Models Skew Western}
Across all tasks, models, and datasets, we find statistically significant moderate correlations with Western, educated, White, and young populations, indicating that language technologies are WEIRD to an extent, though each to varying degrees. 
Prior work identifies Western-centric biases in NLP research~\cite{hershcovich-etal-2022-challenges}, as a majority of research is conducted in the West~\cite{acl_2017, caines_2021}. \citet{joshi-etal-2020-state,blasi-etal-2022-systematic} find disproportionate amounts of resources dedicated to English in NLP research, while \citet{ghosh2021detecting} identify cross-geographic errors made by toxicity models in non-Western contexts. 
This could lead to serious downstream implications such as language extinction~\cite{10.1371/journal.pone.0077056}. Not addressing these biases risks imposing Western standards on non-Western populations, potentially resulting in a new kind of colonialism in the digital age~\cite{irani2010postcolonial}.

\paragraph{Some Populations Are Left Behind}
Certain demographics consistently rank lowest in their alignment with datasets and models across both tasks compared to other demographics of the same type. Prior work has also reported various biases against these populations in datasets and models: people who are non-binary~\cite[e.g.,][]{dev-etal-2021-harms}, Black~\cite[e.g.,][]{sap2019risk,davidson2019racial}, Latinx~\cite[e.g.,][]{dodge-etal-2021-documenting}, Native American~\cite[e.g.,][]{mager-etal-2018-challenges};
and people who are not native English speakers~\cite[e.g.,][]{joshi-etal-2020-state}. These communities are historically marginalized by technological systems~\cite{bender2021dangers}.

\paragraph{Datasets Tend to Align with Their Annotators}
We observe that the positionality we compute is similar to the reported annotator demographics of the datasets, indicating that annotator background contributes to dataset positionality. Social Chemistry reports their annotators largely being women, White, between 30-39 years old, having a college education, and from the U.S.~\cite{forbes-etal-2020-social}, all of which have high correlation to the dataset. Similarly, Dynahate exhibits high correlation with their annotator populations, which are mostly women, White, 18-29 years old, native English speakers, and British~\cite{vidgen-etal-2021-learning}. This could be because annotators' positionalities cause them to make implicit assumptions about the context of subjective annotation tasks, which affects its labels~\cite{wan2023everyone,birhane2022values}. In toxicity modeling, 
men and women value speaking freely versus feeling safe online differently~\cite{duggan2014online}.

\paragraph{Recommendations}
Based on these findings, we discuss some recommendations.
Following prior work on documenting the choices made in building datasets~\cite{gebru2021datasheets} and models~\cite{bender-friedman-2018-data,bender2021dangers}, researchers should keep a record of all design choices made while building them. This can improve reproducibility~\cite{naacl2021reproducibility,aaai2023reproducibility} and aid others in understanding the rationale behind the decisions, revealing some of the researcher's positionality.
Similar to the ``Bender Rule''~\cite{bender2019benderrule}, which suggests stating the language used, researchers should report their positionality and the assumptions they make (potentially after paper acceptance to preserve anonymity).

We echo prior work in recommending methods to center the perspectives of communities who are harmed by design biases~\cite{blodgett2020language, hanna2020towards, bender2021dangers}. This can be done using approaches such as participatory design~\cite{spinuzzi2005methodology}, including interactive storyboarding~\cite{madsen1993experiences}, as well as value-sensitive design~\cite{friedman1996value}, including panels of experiential experts~\cite{madsen1993experiences}. Building datasets and models with large global teams such as BigBench~\cite{srivastava2022beyond} and NL-Augmenter~\cite{dhole2021nl} could also reduce design biases by having diverse teams~\cite{li2020build}.

To account for annotator subjectivity~\cite{aroyo2015truth}, researchers should make concerted efforts to recruit annotators from diverse backgrounds. Websites like  LabintheWild
can be platforms where these annotators are recruited. Since new design biases could be introduced in this process, we recommend following the practice of documenting the demographics of annotators as in prior works \cite[e.g.,][]{forbes-etal-2020-social, vidgen-etal-2021-learning} to record a dataset's positionality. 

We urge considering research through the lens of perspectivism~\cite{basile2021toward}, i.e. being mindful of different perspectives by sharing datasets with disaggregated annotations and finding modeling techniques that can handle inherent disagreements or distributions~\cite{plank2022problem}, instead of forcing a single answer in the data \cite[e.g., by majority vote;][]{davani2022dealing} or model \cite[e.g., by classification to one label;][]{costanza2018design}.
Researchers also should carefully consider how they aggregate labels from diverse annotators during modeling so their perspectives are represented, such as not averaging annotations to avoid the ``tyranny of the mean''~\cite{talat-etal-2022-machine}.

Finally, we argue that the notion of ``inclusive NLP'' does not mean that all language technologies have to work for everyone. Specialized datasets and models are immensely valuable when the data collection process and other design choices are intentional and made to uplift minority voices or historically underrepresented cultures and languages, such as Masakhane-NER~\cite{adelani2021masakhaner} and AfroLM~\cite{dossou2022afrolm}.
There have also been efforts to localize the design of technologies, including applications that adapt their design and functionality to the needs of different cultures~\cite[e.g.,][]{oyibo2016designing,reinecke2011improving,reinecke2013knowing}. Similarly, language models could be made in more culturally adaptive ways, because one size does not fit all \cite{groenwold-etal-2020-investigating,Rettberg2022-tq}. 
Therefore, we urge the NLP community to value the adaptation of language technologies from one language or culture to another~\cite{joshi-etal-2020-state}.

%% file: sections/06_conclusion.tex
\section{Conclusion}
We introduce \framework, a framework to quantify design biases and positionality of datasets and models. In this work, we present how researcher positionality leads to design biases and subsequently gives positionality to datasets and models, potentially resulting in these artifacts not working equally for all populations. Our framework involves recruiting a demographically diverse pool of crowdworkers from around the world on LabintheWild, who then re-annotate a sample of a dataset for an NLP task. We apply \framework to two tasks, social acceptability and hate speech detection, to show that models and datasets have a positionality and design biases by aligning better with Western, White, college-educated, and younger populations. Our results indicate the need for more inclusive models and datasets, paving the way for NLP research that benefits all people.

%% file: sections/98_ethics.tex
\section{Limitations}
\label{sec:limitations}
Our study has several limitations. First, demographics may not be the best construct for positionality, as there may be variability of beliefs within demographic groups. Assuming that there is homogeneity within demographic groups is reductionist and limited. Rather, capturing an individual's attitudes or beliefs may be a more reliable way to capture one's positionality that future work can investigate.

Study annotators could also purposefully answer untruthfully, producing low-quality annotations. We address this risk by using LabintheWild. LabintheWild has been shown to produce high-quality data because participants are intrinsically motivated to participate by learning something about themselves~\cite{reinecke2015labinthewild}. However, as is the case for all online recruiting methods, our sample of participants is not representative of the world's population due to the necessity of having access to the Internet. In addition, there is likely a selection bias in who decides to participate in a LabintheWild study.

Pearson's $r$ may not fully capture alignment as it does not consider interaction effects between different demographics (i.e., intersectionality). Thus, there may be additional mediating or moderating variables that may explain the results that our analysis does not consider. We also took the average of the annotations per group, which could mask individual variations~\cite{talat-etal-2022-machine}. Also, having a low number of participants from specific demographic groups may limit how well the results generalize to the entire group; further, it may risk tokenizing already marginalized communities.
As part of our study, we apply \framework to only two tasks which have relatively straightforward annotation schemes. It may be difficult to generalize to other NLP tasks which have harder annotation schemes, especially ones that require a lot of explanation to the annotators, for example, natural language inference (NLI) tasks.

Our approach is evaluated and works the best for classification tasks and classifiers. Generation tasks would need more careful annotator training which is difficult to achieve on a voluntary platform without adequate incentives. Having annotators use one Likert scale to rate the social acceptability and toxicity of a situation or text may not be a sufficient measure to represent these complex social phenomena. To reduce this threat, we provide detailed instructions that describe how to provide annotations and followed the original annotation setup as closely as possible.

\section{Ethics Statement}\label{sec:ethics}
\paragraph{Towards Inclusive NLP Systems}
Building inclusive NLP systems is important so that everyone can benefit from their usage. Currently, these systems exhibit many design biases that negatively impact minoritized or underserved communities in NLP~\cite{joshi-etal-2020-state,blodgett2020language,bender2021dangers}. Our work is a step towards reducing these disparities by understanding that models and datasets have positionalities and by identifying design biases. The authors take inspiration from fields outside of NLP by studying positionality~\cite{rowe2014positionality} and acknowledge cross-disciplinary research as crucial to building inclusive AI systems.

\paragraph{Ethical Considerations}
We recognize that the demographics we collected only represent a small portion of a person's positionality. There are many aspects of positionality that we did not collect, such as sexual orientation, socioeconomic status, ability, and size. Further, we acknowledge the limitation of assigning labels to people as being inherently reductionist. As mentioned in \S\ref{sec:limitations}, using a single Likert scale for social acceptability and toxicity is not sufficient in capturing the complexities in these phenomena, such as situational context.

We note that quantifying positionality of existing systems is not an endorsement of the system. In addition to making sure that language technologies work for all populations, researchers should also continue to examine whether these systems should exist in the first place~\cite{denton2020tutorial,keyes2019mulching}. Further, we note that understanding a dataset or model's positionality does not preclude researchers from the responsibilities of adjusting it further.

This study was undertaken following approval from the IRB at \redact{the University of Washington} \redact{(STUDY00014813)}. LabintheWild annotators were not compensated financially. They were lay people from a wide range of ages (including minors) and diverse backgrounds. Participants were asked for informed consent to the study procedures as well as the associated risks, such as being exposed to toxic or mature content, prior to beginning the study.

\paragraph{Research Team Positionality}
We discuss aspects of our positionality below that we believe are most relevant to this research.
The research team is comprised of computer scientists who study human-computer interaction and NLP and have a bent for using quantitative methods. Thus,  we approach the topic from a perspective that assumes that positionality can be characterized, fixed, and quantified.

The entire research team currently resides in \redact{the United States}. In alphabetical order, the team members originate from \redact{Belgium and Switzerland}, \redact{France}, \redact{Germany}, \redact{India}, and \redact{the United States}; and identify as \redact{East Asian}, \redact{South Asian}, and \redact{White}. These nationalities and ethnicities are overrepresented in the development of NLP technologies. Thus, we acknowledge that our knowledge of how design biases in NLP datasets and models impact people is largely through research, rather than personal experience.

\newcommand{\icon}[1]{{\includegraphics[height=1.5\fontcharht\font`\B]{#1}}\xspace}
\newcommand{\meiicon}{\icon{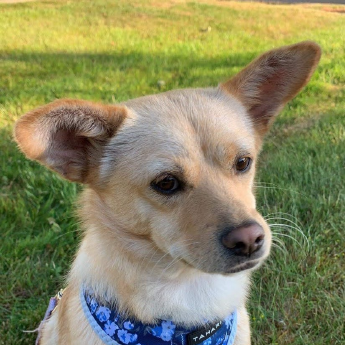}}

\section*{Acknowledgements}
We thank Yejin Choi and Liwei Jiang for their invaluable inputs in the early stages of the project, especially their ideas in shaping the direction of this work, as well as the ReViz team at the Allen Institute for AI for their technical support for building the LabintheWild experiments. We also thank the members of the University of Washington NLP, HCI, and ML/AI groups for their feedback throughout the project. We give a special thanks to Mei \meiicon, an outstanding canine researcher, for providing support and motivation throughout the study.

Jenny T. Liang was supported by the National Science Foundation under grants DGE1745016 and DGE2140739. Katharina Reinecke was partially supported by the National Science Foundation under grant 2230466.

%% file: sections/99_appendix.tex
\clearpage

\appendix

\section{Data}
In this section, we describe all the decisions that went into sampling data points from the different datasets and its post-processing.

\subsection{Sampling}
\label{appendix:sampling}
For Social Chemistry, we sample instances whose label for anticipated agreement by the general public was ``Controversial ($\sim5$0\%)''. We ensure the samples are equally represented by the moral foundation label, which we compute based on majority vote across annotators. In the study, annotators respond whether they found a presented action socially acceptable.

For Dynahate, we randomly sample instances from rounds 3 and 4. In these rounds, annotators generated examples of implicit hate, which is subtler and harder to detect and could yield differences in annotations. We ensure that there are equal amounts of hateful and not hateful instances and that the types of targets of the hateful instances are equally represented. During the studsy, annotators respond whether they found a presented instance toxic.

For both social acceptability and hate speech detection, annotators responded whether they found the situation moral and whether they found the instance to be hate speech respectively. 

\subsection{Post-Processing}
Because Social Chemistry has multiple annotations for each instance, we compute an aggregate score by taking the average score across annotators. This score is then used to correlate to the annotators' aggregated scores.

\section{Study Design}
In this section, we discuss the design of the LabintheWild experiments. The social acceptability task was released to the public in April 2022. The hate speech detection task was released August 2022. To reduce confounding factors on the data collection process, we conduct multiple user studies of the LabintheWild experiments prior to the public release. Additionally, all the annotations collected through the experiments are anonymous and are stored securely.

The social acceptability task is marketed as ``Could you live with an AI and its morals?'' Participants for this study provide annotations for 25 situations. The hate speech detection task is marketed as ``Do you and AI agree on what is hate speech? Let's find out!'' Participants provid annotations for 15 instances.  

\subsection{LabintheWild Study Flow}
We describe the format of the LabintheWild experiment. The phases of the experiment are: obtaining consent, collecting demographics, explaining instructions, collecting annotations, collecting study feedback, and displaying results.

\paragraph{Obtaining Consent}
Prior to beginning the study, participants review a consent form. The consent form includes information on the purpose of the research, what the participant will do, risks and benefits of the research, privacy and data collection methods, and contact information of the researchers. At the end of the form, participants give explicit consent to participate in the study.

\paragraph{Collecting Demographics}
We then collect the demographics of study participants. LabintheWild participants enter in whether they had taken this test before, the country they lived in the longest, the country of residence, age, native language, religion, education, and ethnicity. No demographics are required except for the country the participant lived in the longest and whether they have taken the test before. Additionally, we only display ethnicity for people within the United States.

\begin{figure*}
    \centering
    \includegraphics[width=0.9\textwidth]{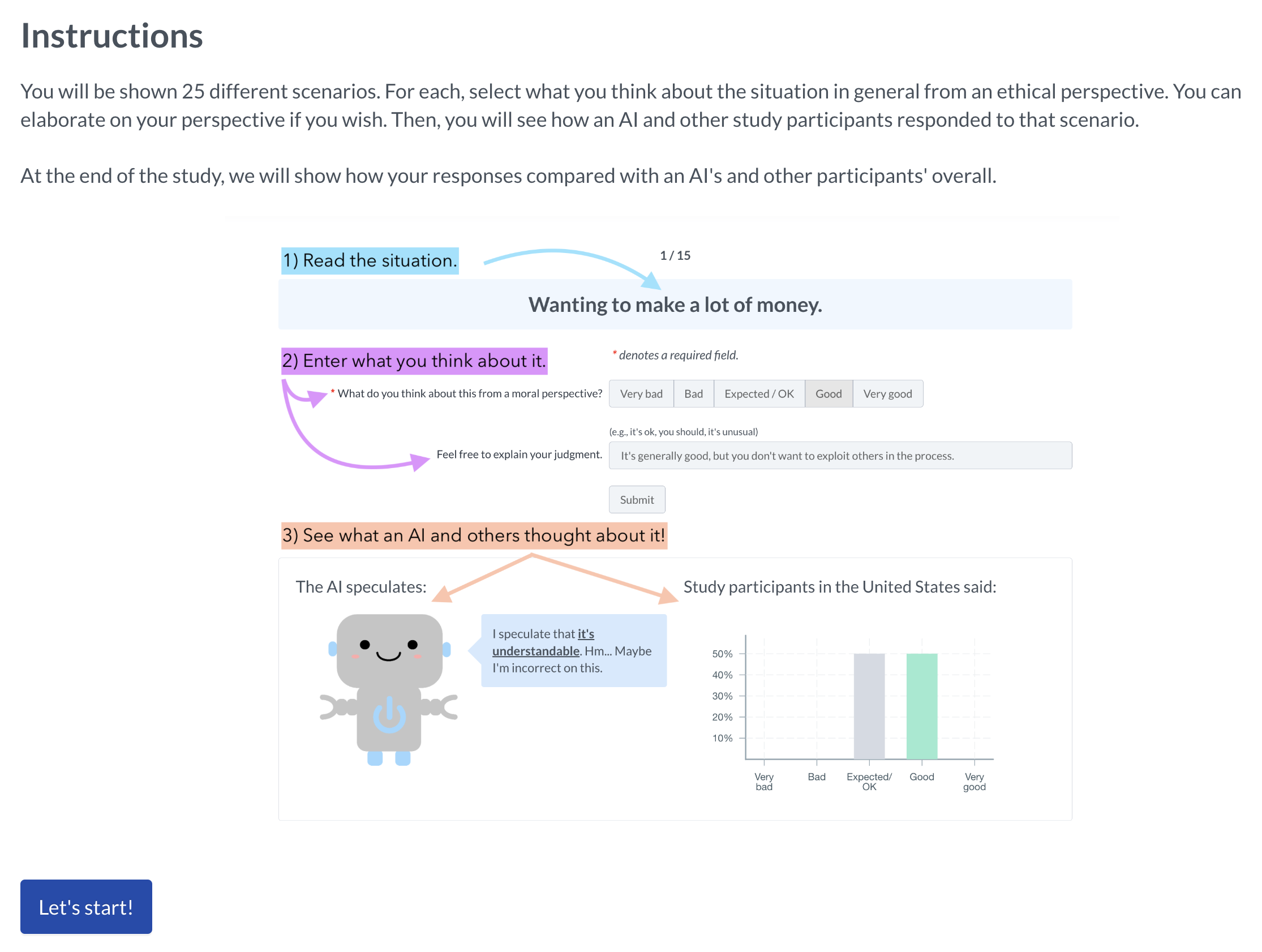}
    \caption{\textbf{Instructions for the social acceptability task.} Participants were asked to describe their thoughts about a situation from an ethical perspective.}
    \label{fig:acceptability_instructions}
\end{figure*}

\begin{figure*}
    \centering
    \includegraphics[width=0.9\textwidth]{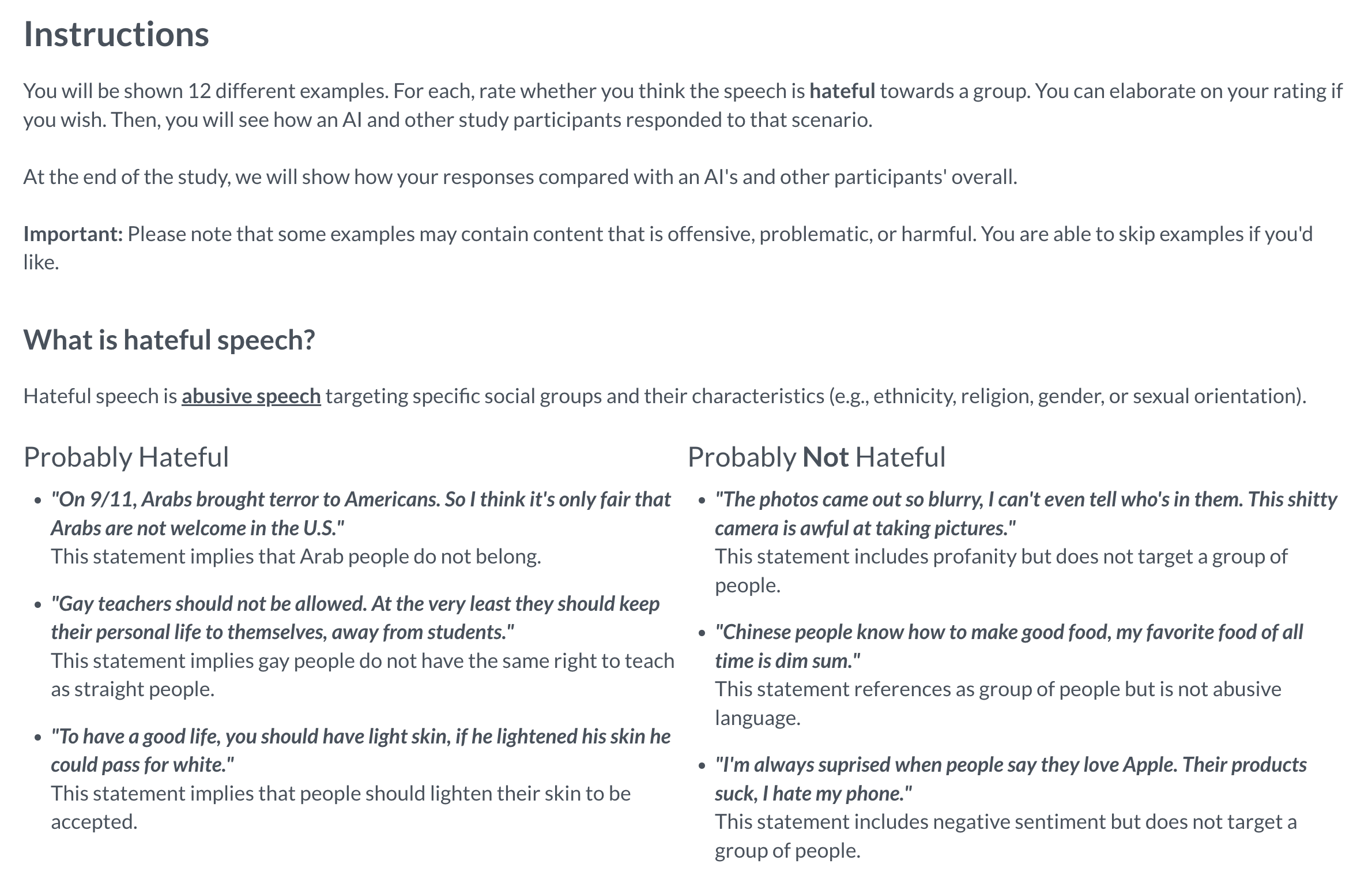}
    \caption{\textbf{Instructions for the toxicity task.} Participants were provided with examples of hate speech examples and not hate examples.}
    \label{fig:toxicity_instructions}
\end{figure*}

\paragraph{Explaining Instructions}
For each task, we provide instructions to participants on how to perform the annotation task. For social acceptability, we explain social acceptability as rating ``what you think about the situation in general from an ethical perspective'' (see Figure~\ref{fig:acceptability_instructions}). For hate speech detection, use the definition of hate speech from Dynahate and we provide three examples of hate speech (see Figure~\ref{fig:toxicity_instructions}). We also present examples of negative sentiment, profanity, or discussing groups that could be confused as hate speech, but are not hate speech.

\begin{figure*}
    \centering
    \includegraphics[width=0.9\textwidth]{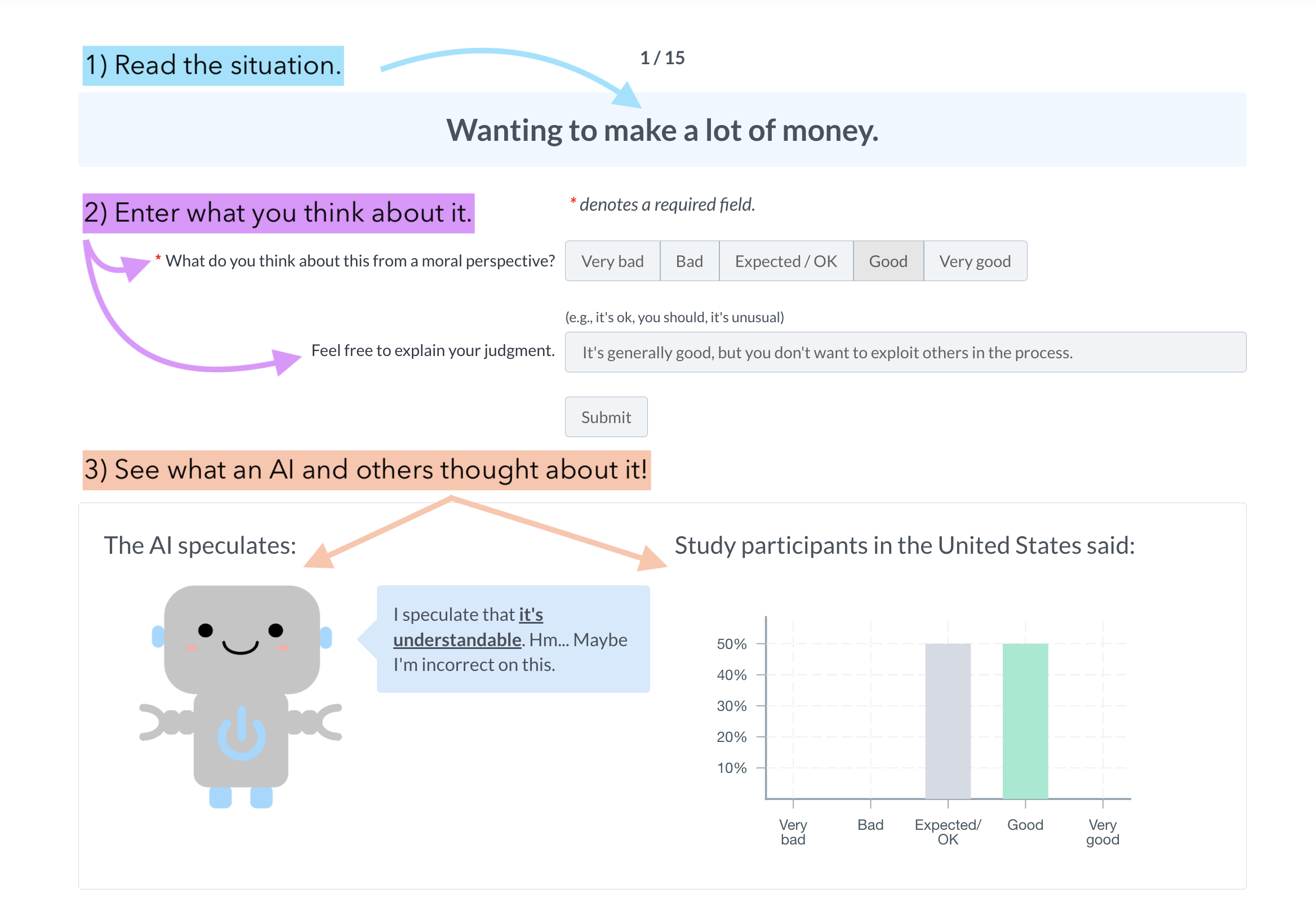}
    \caption{\textbf{Data collection interface for the social acceptability task.} Participants were given a sentence (an action from the Social Chemsitry dataset) and asked to rate how ethical the action was. Participants are shown how other people from their country responded after each attempt.}
    \label{fig:acceptability_interface}
\end{figure*}

\begin{figure*}
    \centering
    \includegraphics[width=0.9\textwidth]{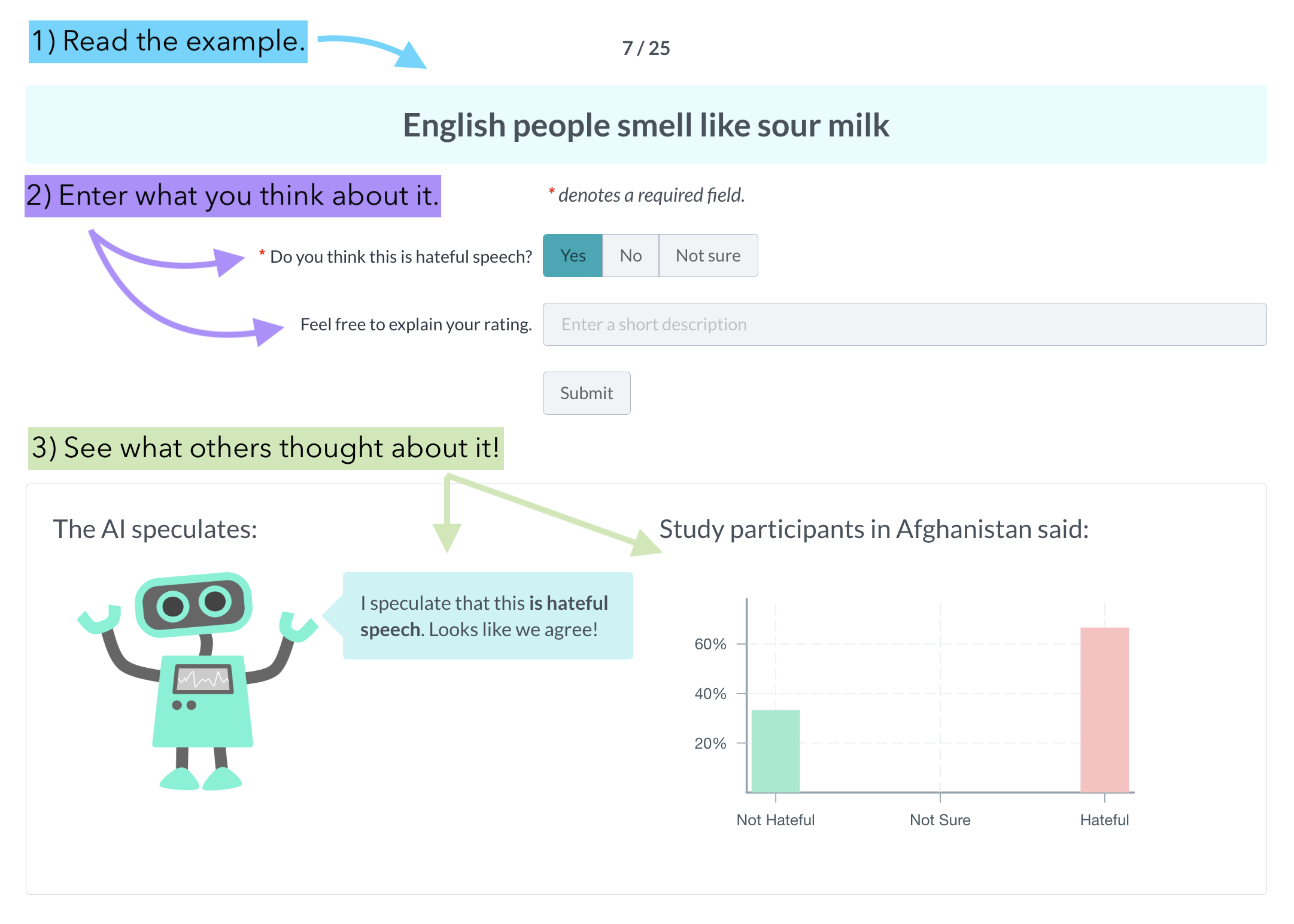}
    \caption{\textbf{Data collection interface for the hate speech task.} Participants were given a sentence (an instance from the Dynahate dataset) and asked to rate whether the instance was toxic or not. Participants are shown how other people from their country responded after each attempt.}
    \label{fig:toxicity_interface}
\end{figure*}

\paragraph{Collecting Annotations}
After being presented with instructions, participants begin data collection from the 300 instances selected from Section~\ref{appendix:sampling}. For each task, we keep the annotation setup identical to the original one. For social acceptability, we collect Likert-scale ratings of situations ranging from ``It's very bad'', ``It's bad'', ``It's okay'', ``It's good'', and ``It's very good''. Participants can provide rationale for their decision by using an open text box. The data collection interface is presented in Figure~\ref{fig:acceptability_interface}. For hate speech detection, we collect ratings of instances ranging from ``Hate speech'', ``Not sure'', ``Not hate speech''. We also provide an optional open-text box for participants to explain their rationale. The data collection interface is presented in Figure~\ref{fig:toxicity_interface}. After submitting the annotation, the participant is able to see a visualization on how the AI responded as well as how other participants from the same country responded to the instance.

We also specifically sample which instances to present to participants for annotation. We sample a third of the instances that did not have any annotations from the demographic and a third that are already sampled by participants of the demographic. The rest are equally split across the different of types of instances (i.e., moral foundation for Social Chemistry, hate type for Dynahate).

\paragraph{Providing Study Feedback}
Following typical LabintheWild experiment procedures, we collect feedback from participants about the study. Participants can enter open-text feedback on anything. They also submit whether they encountered technical difficulties during the study or whether they cheated. Participants can elaborate on their answers from the prior questions in an open-text box.

\begin{figure*}
    \centering
    \includegraphics[width=0.9\textwidth]
    {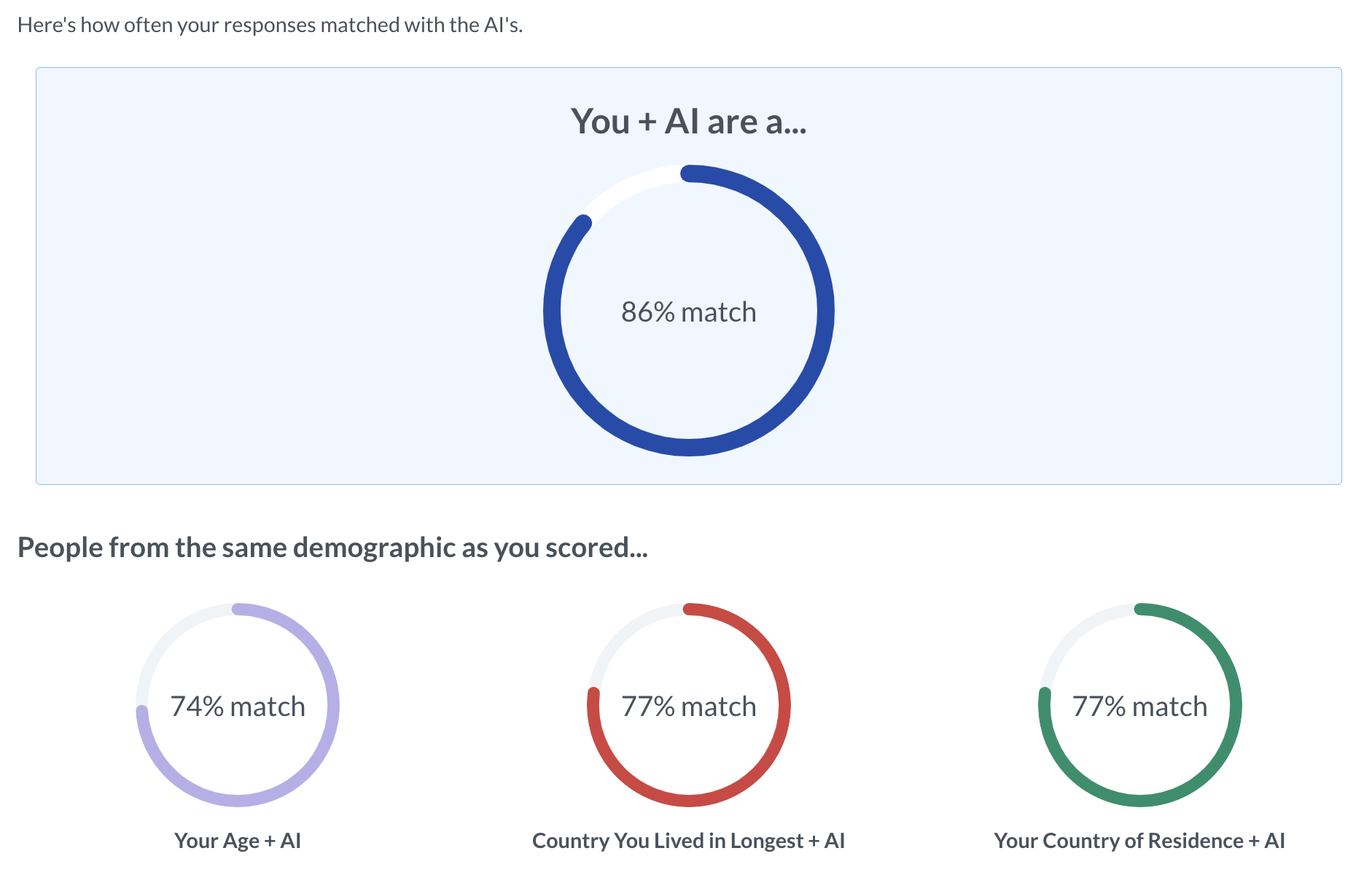}
    \vspace{1em}
    \includegraphics[width=0.9\textwidth]{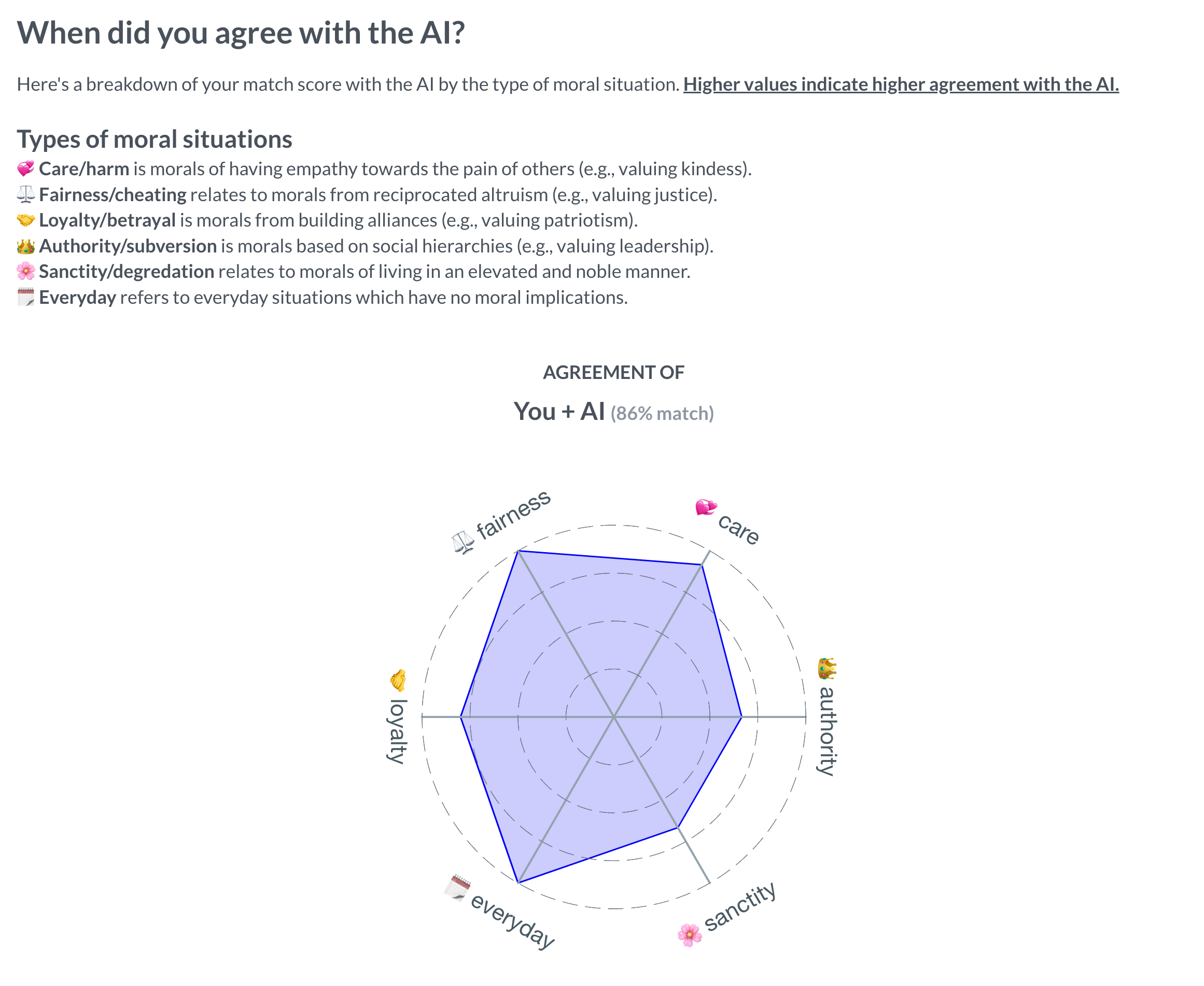}
    \caption{\textbf{Results interface for the social acceptability task.} Participants can view how well they aligned with the AI, as well as how other demographics they reported aligned with the AI. The AI alignment is further broken down by the type of moral foundation.}
    \label{fig:results-interface}
\end{figure*}

\paragraph{Displaying Overall Results}
Finally, participants see their overall results for the experiment task. First, participants are presented with the percentage of time they agreed with the AI as well as with participants as the same demographic as them (see Figure~\ref{fig:results-interface}). Each of these agreement scores are further broken down by the type of the instance (i.e., moral foundation for Social Chemistry and hate type for Dynahate).

\label{appendix:study-design}

\section{Additional Results}
In this section, we report additional results from our analyses of the LabintheWild data.

\subsection{\emph{p}-values}
\label{sec:p-values}
We report the \emph{p}-values from our analyses from Table~\ref{tab:pvalues}.

\input{tables/p-values}

\section{Cultural Spheres}
Division of countries can be done through continents. However, continents are often not representative of the countries within it and clustering based on them can lead to inaccurate findings. For example, Asia includes both Japan and Saudi Arabia, which are different culturally. We instead adopt cultural spheres as used in World Values Survey~\cite{haerpfer2012world}, which clusters the countries in terms of the values they uphold and norms they follow. Table \ref{tab:culturalspheres} shows the countries and the spheres.

\begin{table*}
\small
    \centering
    \begin{tabular}{p{3cm}p{12cm}}
    \toprule
     \textbf{Cultural Sphere} &  \textbf{Countries}\\
    \midrule
     African-Islamic & 
        Afghanistan, Albania, Algeria, Azerbaijan, Ethiopia, Indonesia, Iraq, Jordan, Morocco, Pakistan, Palestine, Qatar, Nigeria, Saudi Arabia, South Africa, Syrian Arab Republic, Tunisia, Turkey, United Arab Emirates, Uzbekistan
        \newline
        \textcolor{gray!50}{Burkina Faso, Bangladesh, Egypt, Ghana, Iran, Kazakhstan, Kyrgyzstan, Lebanon, Libya, Mali, Rwanda, Tajikistan, Tanzania, Uganda, Yemen, Zambia, Zimbabwe}
        \vspace{\spsize}\\
        
     Baltic & 
        Estonia, Latvia, Lithuania, Åland Islands
        \vspace{\spsize}\\

     Catholic-Europe & 
       Andorra, Austria, Belgium, Czech Republic, France, Hungary, Italy, Luxembourg, Poland, Portugal, Spain
       \newline
       \textcolor{gray!50}{Slovakia, Slovenia}
       \vspace{\spsize}\\
       
     Confucian &
        China, Hong Kong, Japan, South Korea, Taiwan
        \newline
        \textcolor{gray!50}{Macao}
        \vspace{\spsize}\\
        
     English-Speaking &
        American Samoa, Australia, Canada, Guernsey, Ireland, New Zealand, United Kingdom, United States
        \vspace{\spsize}\\
        
     Latin-America & 
        Argentina, Brazil, Colombia, Dominican Republic, Mexico, Philippines, Trinidad and Tobago, Venezuela
        \newline
        \textcolor{gray!50}{Bolivia, Chile, Ecuador, Guatemala, Haiti, Nicaragua, Peru, Puerto Rico, Uruguay}
        \vspace{\spsize}\\

     Orthodox-Europe &
        Belarus, Bosnia, Bulgaria, Cyprus, Georgia, Greece, Moldova, Romania, Russia, Serbia, Ukraine
        \newline
        \textcolor{gray!50}{Armenia, Montenegro, North Macedonia}
        \vspace{\spsize}\\

     Protestant-Europe &
        Denmark, Finland, Germany, Iceland, Netherlands, Norway, Sweden, Switzerland
        \vspace{\spsize}\\

    West-South-Asia &
        India, Israel, Malaysia, Myanmar, Singapore, Vietnam
        \newline
        \textcolor{gray!50}{Thailand}
        \vspace{\spsize}\\

    \bottomrule
    \end{tabular}
    \caption{\textbf{Cultural spheres and their corresponding countries from~\cite{haerpfer2012world}.} Black color indicates that the countries are part of our collected data. Gray color indicates countries not part of our analysis---we have included them to give an idea of which other countries belong to the spheres.}
    \label{tab:culturalspheres}
\end{table*}

%% file: tables/p-values.tex
\begin{table*}[]
\scriptsize
    \centering
    \begin{tabular}{p{5cm}">{\columncolor{datasetcolor}}P{0.87cm}P{0.87cm}P{0.87cm}">{\columncolor{datasetcolor}}P{0.87cm}P{0.87cm}P{0.87cm}P{0.87cm}P{0.87cm}"}
    \toprule
    \multicolumn{9}{l}{
    {\fontfamily{lmtt}\selectfont
     DATASETS:} {\scriptsize \socialchemlogo SocialChemistry \quad \dynahatelogo DynaHate} \hfill {\fontfamily{lmtt}\selectfont MODELS:} {\scriptsize \gptlogo GPT-4 \quad \delphilogo Delphi \quad \perspectivelogo PerspectiveAPI \quad \rewirelogo RewireAPI \quad \hatebertlogo HateRoberta}} \\
    \toprule
      \textbf{Demographic} &  \multicolumn{8}{c"}{\textbf{$p$-value} ($\alpha$ = 2.04e-05)}\\
    \midrule
    & \multicolumn{3}{c"}{\ul{\textbf{Social Acceptability}}} & \multicolumn{5}{c"}{\ul{\textbf{Toxicity \& Hate Speech}}}\\[\tabsp]
     & \socialchemlogo & \delphilogo & \gptlogo & \dynahatelogo & \perspectivelogo & \rewirelogo & \hatebertlogo & \gptlogo \\
    \cmidrule{2-9}
\sethlcolor{countrycolor} 
 \textbf{\codebox{} Country (Lived Longest)} \dotfilla & \multicolumn{8}{c"}{\dotfilla\dotfilla\dotfilla\dotfilla\dotfilla\dotfilla\dotfilla\dotfilla\dotfilla\dotfilla\dotfilla\dotfilla\dotfilla\dotfilla\dotfilla\dotfilla\dotfilla\dotfilla\dotfilla\dotfilla\dotfilla\dotfilla\dotfilla\dotfilla\dotfilla\dotfilla\dotfilla\dotfilla\dotfilla\dotfilla\dotfilla\dotfilla\dotfilla\dotfilla\dotfilla\dotfilla\dotfilla\dotfilla\dotfilla\dotfilla\dotfilla\dotfilla\dotfilla\dotfilla\dotfilla\dotfilla\dotfilla\dotfilla\dotfilla\dotfilla\dotfilla\dotfilla\dotfilla\dotfilla\dotfilla\dotfilla\dotfilla\dotfilla\dotfilla\dotfilla\dotfilla\dotfilla\dotfilla\dotfilla\dotfilla\dotfilla\dotfilla\dotfilla\dotfilla\dotfilla\dotfilla\dotfilla\dotfilla\dotfilla\dotfilla\dotfilla\dotfilla} \\ 
\quad African Islamic                     & 1.74e-04 & 2.01e-03 & 4.40e-03  & 4.02e-03 & 2.37e-01 & 3.50e-03  & 3.28e-01 & 6.82e-01 \\
\quad Baltic                              & 2.98e-06 & 7.11e-06 & 1.27e-05 & 1.00e+00    & 1.00e+00    & 1.00e+00    & 1.00e+00    & 4.34e-01 \\
\quad Catholic Europe                     & 1.40e-09  & 1.98e-07 & 3.77e-11 & 2.21e-01 & 1.00e+00    & 2.01e-01 & 1.00e+00    & 1.00e+00    \\
\quad Confucian                           & 5.23e-15 & 3.89e-07 & 1.58e-14 & 3.15e-03 & 1.00e+00    & 4.27e-04 & 1.00e+00    & 3.07e-04 \\
\quad English-Speaking                    & 6.67e-55 & 4.12e-29 & 2.21e-49 & 3.31e-44 & 3.59e-07 & 8.74e-27 & 3.17e-09 & 5.38e-12 \\
\quad Latin American                      & 2.50e-02  & 9.08e-02 & 1.52e-02 & 7.87e-01 & 1.00e+00    & 1.00e+00    & 1.00e+00    & 1.00e+00    \\
\quad Orthodox Europe                     & 1.02e-06 & 2.42e-07 & 1.38e-10 & 1.37e-01 & 1.00e+00    & 3.34e-03 & 1.00e+00    & 1.00e+00    \\
\quad Protestant Europe                   & 1.17e-14 & 2.18e-10 & 6.14e-16 & 1.15e-04 & 1.46e-02 & 5.09e-01 & 5.43e-02 & 5.07e-03 \\
\quad West South Asia                     & 1.63e-09 & 2.10e-08  & 4.53e-08 & 3.30e-01  & 1.00e+00    & 4.34e-01 & 9.13e-01 & 1.00e+00    \\
\sethlcolor{educationcolor} 
 \textbf{\codebox{} Education Level} \dotfilla & \multicolumn{8}{c"}{\dotfilla\dotfilla\dotfilla\dotfilla\dotfilla\dotfilla\dotfilla\dotfilla\dotfilla\dotfilla\dotfilla\dotfilla\dotfilla\dotfilla\dotfilla\dotfilla\dotfilla\dotfilla\dotfilla\dotfilla\dotfilla\dotfilla\dotfilla\dotfilla\dotfilla\dotfilla\dotfilla\dotfilla\dotfilla\dotfilla\dotfilla\dotfilla\dotfilla\dotfilla\dotfilla\dotfilla\dotfilla\dotfilla\dotfilla\dotfilla\dotfilla\dotfilla\dotfilla\dotfilla\dotfilla\dotfilla\dotfilla\dotfilla\dotfilla\dotfilla\dotfilla\dotfilla\dotfilla\dotfilla\dotfilla\dotfilla\dotfilla\dotfilla\dotfilla\dotfilla\dotfilla\dotfilla\dotfilla\dotfilla\dotfilla\dotfilla\dotfilla\dotfilla\dotfilla\dotfilla\dotfilla\dotfilla\dotfilla\dotfilla\dotfilla\dotfilla\dotfilla} \\ 
\quad College                             & 1.02e-50 & 1.19e-35 & 8.21e-41 & 8.96e-37 & 8.42e-08 & 7.75e-25 & 9.17e-10 & 8.75e-11 \\
\quad Graduate School                     & 5.80e-44  & 1.97e-21 & 1.74e-39 & 9.60e-23  & 3.79e-04 & 4.51e-16 & 3.15e-03 & 4.12e-08 \\
\quad High School                         & 9.32e-38 & 1.31e-21 & 4.85e-33 & 6.01e-24 & 2.74e-03 & 1.19e-14 & 4.48e-05 & 5.12e-08 \\
\quad PhD                                 & 4.16e-28 & 2.29e-18 & 4.32e-24 & 1.63e-09 & 5.54e-01 & 9.82e-08 & 2.54e-02 & 1.93e-03 \\
\quad Pre-High School                     & 4.48e-17 & 8.53e-11 & 7.00e-20    & 2.25e-02 & 1.00e+00    & 8.06e-04 & 1.00e+00    & 1.43e-02 \\
\quad Professional School                 & 2.19e-13 & 1.50e-09  & 3.50e-11  & 1.65e-12 & 1.00e+00    & 3.08e-03 & 1.00e+00    & 1.00e+00    \\
\sethlcolor{ethnicitycolor} 
 \textbf{\codebox{} Ethnicity} \dotfilla & \multicolumn{8}{c"}{\dotfilla\dotfilla\dotfilla\dotfilla\dotfilla\dotfilla\dotfilla\dotfilla\dotfilla\dotfilla\dotfilla\dotfilla\dotfilla\dotfilla\dotfilla\dotfilla\dotfilla\dotfilla\dotfilla\dotfilla\dotfilla\dotfilla\dotfilla\dotfilla\dotfilla\dotfilla\dotfilla\dotfilla\dotfilla\dotfilla\dotfilla\dotfilla\dotfilla\dotfilla\dotfilla\dotfilla\dotfilla\dotfilla\dotfilla\dotfilla\dotfilla\dotfilla\dotfilla\dotfilla\dotfilla\dotfilla\dotfilla\dotfilla\dotfilla\dotfilla\dotfilla\dotfilla\dotfilla\dotfilla\dotfilla\dotfilla\dotfilla\dotfilla\dotfilla\dotfilla\dotfilla\dotfilla\dotfilla\dotfilla\dotfilla\dotfilla\dotfilla\dotfilla\dotfilla\dotfilla\dotfilla\dotfilla\dotfilla\dotfilla\dotfilla\dotfilla\dotfilla} \\ 
\quad Asian, Asian American               & 6.37e-35 & 2.04e-22 & 4.77e-31 & 1.85e-21 & 4.80e-07  & 1.46e-13 & 4.19e-06 & 9.54e-09 \\
\quad Black, African American             & 3.50e-24  & 8.08e-15 & 2.03e-20 & 8.82e-14 & 1.01e-03 & 6.16e-05 & 1.79e-03 & 2.34e-05 \\
\quad Latino / Latina, Hispanic           & 1.47e-19 & 8.00e-13    & 6.30e-14  & 6.39e-07 & 2.39e-05 & 5.23e-08 & 3.19e-03 & 3.26e-03 \\
\quad Native American, Alaskan Native     & 2.33e-07 & 3.11e-05 & 3.44e-09 & 1.00e+00    & 6.37e-01 & 6.72e-01 & 6.07e-01 & 4.81e-01 \\
\quad Pacific Islander, Native Australian & 6.63e-04 & 1.38e-03 & 2.22e-03 & 1.00e+00    & 1.32e-02 & 1.77e-01 & 1.59e-02 & 1.01e-01 \\
\quad White                               & 1.27e-48 & 4.94e-29 & 1.44e-42 & 4.51e-42 & 1.47e-05 & 2.00e-24    & 1.18e-06 & 8.31e-10 \\
\sethlcolor{gendercolor} 
 \textbf{\codebox{} Gender} \dotfilla & \multicolumn{8}{c"}{\dotfilla\dotfilla\dotfilla\dotfilla\dotfilla\dotfilla\dotfilla\dotfilla\dotfilla\dotfilla\dotfilla\dotfilla\dotfilla\dotfilla\dotfilla\dotfilla\dotfilla\dotfilla\dotfilla\dotfilla\dotfilla\dotfilla\dotfilla\dotfilla\dotfilla\dotfilla\dotfilla\dotfilla\dotfilla\dotfilla\dotfilla\dotfilla\dotfilla\dotfilla\dotfilla\dotfilla\dotfilla\dotfilla\dotfilla\dotfilla\dotfilla\dotfilla\dotfilla\dotfilla\dotfilla\dotfilla\dotfilla\dotfilla\dotfilla\dotfilla\dotfilla\dotfilla\dotfilla\dotfilla\dotfilla\dotfilla\dotfilla\dotfilla\dotfilla\dotfilla\dotfilla\dotfilla\dotfilla\dotfilla\dotfilla\dotfilla\dotfilla\dotfilla\dotfilla\dotfilla\dotfilla\dotfilla\dotfilla\dotfilla\dotfilla\dotfilla\dotfilla} \\ 
\quad Man                                 & 2.55e-47 & 2.19e-31 & 8.72e-41 & 1.99e-34 & 1.09e-07 & 3.55e-24 & 7.84e-08 & 1.46e-08 \\
\quad Non-Binary                          & 3.61e-26 & 4.94e-18 & 1.14e-21 & 3.00e-16    & 1.64e-01 & 6.67e-06 & 8.00e-03    & 8.49e-04 \\
\quad Woman                               & 7.04e-51 & 1.25e-27 & 1.76e-48 & 4.02e-33 & 6.36e-08 & 8.19e-22 & 4.27e-10 & 2.17e-09 \\
\sethlcolor{languagecolor} 
 \textbf{\codebox{} Native Language} \dotfilla & \multicolumn{8}{c"}{\dotfilla\dotfilla\dotfilla\dotfilla\dotfilla\dotfilla\dotfilla\dotfilla\dotfilla\dotfilla\dotfilla\dotfilla\dotfilla\dotfilla\dotfilla\dotfilla\dotfilla\dotfilla\dotfilla\dotfilla\dotfilla\dotfilla\dotfilla\dotfilla\dotfilla\dotfilla\dotfilla\dotfilla\dotfilla\dotfilla\dotfilla\dotfilla\dotfilla\dotfilla\dotfilla\dotfilla\dotfilla\dotfilla\dotfilla\dotfilla\dotfilla\dotfilla\dotfilla\dotfilla\dotfilla\dotfilla\dotfilla\dotfilla\dotfilla\dotfilla\dotfilla\dotfilla\dotfilla\dotfilla\dotfilla\dotfilla\dotfilla\dotfilla\dotfilla\dotfilla\dotfilla\dotfilla\dotfilla\dotfilla\dotfilla\dotfilla\dotfilla\dotfilla\dotfilla\dotfilla\dotfilla\dotfilla\dotfilla\dotfilla\dotfilla\dotfilla\dotfilla} \\ 
\quad English                             & 8.54e-55 & 2.04e-33 & 1.91e-44 & 1.22e-44 & 3.38e-07 & 1.28e-29 & 2.10e-10  & 2.39e-12 \\
\quad Not English                         & 1.04e-25 & 5.10e-18  & 1.05e-27 & 9.78e-11 & 1.58e-04 & 2.40e-07  & 1.93e-04 & 6.29e-06 \\
\sethlcolor{agecolor} 
 \textbf{\codebox{} Age} \dotfilla & \multicolumn{8}{c"}{\dotfilla\dotfilla\dotfilla\dotfilla\dotfilla\dotfilla\dotfilla\dotfilla\dotfilla\dotfilla\dotfilla\dotfilla\dotfilla\dotfilla\dotfilla\dotfilla\dotfilla\dotfilla\dotfilla\dotfilla\dotfilla\dotfilla\dotfilla\dotfilla\dotfilla\dotfilla\dotfilla\dotfilla\dotfilla\dotfilla\dotfilla\dotfilla\dotfilla\dotfilla\dotfilla\dotfilla\dotfilla\dotfilla\dotfilla\dotfilla\dotfilla\dotfilla\dotfilla\dotfilla\dotfilla\dotfilla\dotfilla\dotfilla\dotfilla\dotfilla\dotfilla\dotfilla\dotfilla\dotfilla\dotfilla\dotfilla\dotfilla\dotfilla\dotfilla\dotfilla\dotfilla\dotfilla\dotfilla\dotfilla\dotfilla\dotfilla\dotfilla\dotfilla\dotfilla\dotfilla\dotfilla\dotfilla\dotfilla\dotfilla\dotfilla\dotfilla\dotfilla} \\ 
\quad 10-20 yrs old                       & 5.54e-43 & 9.00e-29    & 1.46e-40 & 2.89e-29 & 1.85e-06 & 2.23e-22 & 7.63e-09 & 8.33e-09 \\
\quad 20-30 yrs old                       & 5.35e-50 & 1.49e-36 & 1.23e-42 & 1.79e-34 & 1.22e-07 & 6.51e-24 & 5.61e-10 & 2.90e-12 \\
\quad 30-40 yrs old                       & 2.71e-33 & 2.24e-18 & 7.56e-27 & 2.25e-10 & 1.00e+00    & 2.37e-07 & 4.49e-02 & 3.21e-03 \\
\quad 40-50 yrs old                       & 2.48e-24 & 4.36e-18 & 2.98e-26 & 3.43e-16 & 1.49e-02 & 2.12e-12 & 5.43e-03 & 1.68e-04 \\
\quad 50-60 yrs old                       & 9.40e-23  & 9.98e-12 & 4.58e-16 & 1.96e-10 & 1.49e-01 & 9.98e-05 & 1.00e+00    & 2.47e-01 \\
\quad 60-70 yrs old                       & 4.85e-17 & 9.35e-09 & 1.92e-14 & 4.99e-01 & 1.00e+00    & 1.00e+00    & 1.00e+00    & 1.00e+00    \\
\quad 70-80 yrs old                       & 5.14e-05 & 4.20e-04  & 3.91e-05 & 8.78e-01 & 1.00e+00    & 1.00e+00    & 1.00e+00    & 2.96e-05 \\
\quad 80+ yrs old                         & 4.75e-01 & 9.08e-01 & 8.63e-02 & 1.00e+00    & 1.00e+00    & 1.00e+00    & 1.00e+00    & 1.00e+00    \\
\sethlcolor{countrycolor} 
 \textbf{\codebox{} Country (Residence)} \dotfilla & \multicolumn{8}{c"}{\dotfilla\dotfilla\dotfilla\dotfilla\dotfilla\dotfilla\dotfilla\dotfilla\dotfilla\dotfilla\dotfilla\dotfilla\dotfilla\dotfilla\dotfilla\dotfilla\dotfilla\dotfilla\dotfilla\dotfilla\dotfilla\dotfilla\dotfilla\dotfilla\dotfilla\dotfilla\dotfilla\dotfilla\dotfilla\dotfilla\dotfilla\dotfilla\dotfilla\dotfilla\dotfilla\dotfilla\dotfilla\dotfilla\dotfilla\dotfilla\dotfilla\dotfilla\dotfilla\dotfilla\dotfilla\dotfilla\dotfilla\dotfilla\dotfilla\dotfilla\dotfilla\dotfilla\dotfilla\dotfilla\dotfilla\dotfilla\dotfilla\dotfilla\dotfilla\dotfilla\dotfilla\dotfilla\dotfilla\dotfilla\dotfilla\dotfilla\dotfilla\dotfilla\dotfilla\dotfilla\dotfilla\dotfilla\dotfilla\dotfilla\dotfilla\dotfilla\dotfilla} \\ 
\quad African Islamic                     & 2.01e-02 & 2.64e-02 & 4.28e-02 & 2.75e-01 & 1.00e+00    & 1.00e+00    & 1.00e+00    & 1.00e+00    \\
\quad Baltic                              & 8.25e-03 & 8.25e-03 & 1.00e+00    & 1.00e+00    & 1.00e+00    & 1.00e+00    & 1.00e+00    & 1.66e-01 \\
\quad Catholic Europe                     & 6.35e-08 & 3.01e-04 & 7.84e-13 & 1.68e-01 & 1.00e+00    & 1.82e-02 & 1.00e+00    & 1.00e+00    \\
\quad Confucian                           & 3.36e-08 & 1.83e-04 & 1.35e-11 & 1.62e-01 & 4.59e-01 & 5.03e-02 & 8.55e-01 & 2.13e-02 \\
\quad English-Speaking                    & 1.96e-53 & 8.43e-35 & 6.34e-48 & 7.43e-47 & 1.17e-07 & 2.65e-29 & 3.29e-10 & 6.96e-13 \\
\quad Latin American                      & 1.14e-04 & 5.20e-05  & 7.76e-06 & 1.00e+00    & 1.00e+00    & 1.00e+00    & 1.00e+00    & 1.00e+00    \\
\quad Orthodox Europe                     & 2.23e-03 & 1.60e-05  & 3.18e-06 & 1.00e+00    & 1.00e+00    & 4.34e-01 & 1.00e+00    & 1.00e+00    \\
\quad Protestant Europe                   & 6.59e-18 & 5.21e-14 & 3.82e-16 & 3.23e-06 & 1.43e-02 & 3.54e-01 & 1.66e-02 & 1.21e-02  \\
\quad West South Asia                     & 3.46e-08 & 8.91e-07 & 1.29e-05 & 1.89e-03 & 1.00e+00    & 3.46e-01 & 1.00e+00    & 1.00e+00    \\
\sethlcolor{religioncolor} 
 \textbf{\codebox{} Religion} \dotfilla & \multicolumn{8}{c"}{\dotfilla\dotfilla\dotfilla\dotfilla\dotfilla\dotfilla\dotfilla\dotfilla\dotfilla\dotfilla\dotfilla\dotfilla\dotfilla\dotfilla\dotfilla\dotfilla\dotfilla\dotfilla\dotfilla\dotfilla\dotfilla\dotfilla\dotfilla\dotfilla\dotfilla\dotfilla\dotfilla\dotfilla\dotfilla\dotfilla\dotfilla\dotfilla\dotfilla\dotfilla\dotfilla\dotfilla\dotfilla\dotfilla\dotfilla\dotfilla\dotfilla\dotfilla\dotfilla\dotfilla\dotfilla\dotfilla\dotfilla\dotfilla\dotfilla\dotfilla\dotfilla\dotfilla\dotfilla\dotfilla\dotfilla\dotfilla\dotfilla\dotfilla\dotfilla\dotfilla\dotfilla\dotfilla\dotfilla\dotfilla\dotfilla\dotfilla\dotfilla\dotfilla\dotfilla\dotfilla\dotfilla\dotfilla\dotfilla\dotfilla\dotfilla\dotfilla\dotfilla} \\ 
\quad Buddhist                            & 7.42e-13 & 3.16e-10 & 7.78e-09 & 2.44e-02 & 1.00e+00    & 1.00e+00    & 1.00e+00    & 1.27e-02 \\
\quad Christian                           & 3.47e-48 & 2.43e-22 & 9.04e-47 & 1.21e-22 & 1.66e-07 & 3.99e-17 & 3.03e-08 & 3.61e-07 \\
\quad Hindu                               & 4.62e-14 & 3.57e-11 & 2.97e-10 & 1.12e-08 & 7.96e-02 & 6.02e-03 & 3.03e-01 & 1.89e-02 \\
\quad Jewish                              & 8.32e-17 & 1.85e-13 & 4.97e-13 & 8.13e-11 & 1.95e-01 & 4.75e-04 & 1.89e-01 & 4.87e-02 \\
\quad Muslim                              & 2.72e-14 & 1.81e-12 & 1.37e-20 & 7.50e-02  & 1.00e+00    & 1.00e+00    & 1.00e+00    & 1.00e+00    \\
\quad Spritual                            & 9.75e-08 & 3.49e-07 & 3.56e-12 & 1.00e+00    & 1.00e+00    & 1.00e+00    & 1.00e+00    & --- \\
    \bottomrule
    \end{tabular}
    \caption{\textbf{Associated \emph{p}-values of each associated Pearson's $r$ correlation value after applying Bonferroni corrections.} $\alpha$ = 0.001 and $\alpha$ = 2.04e-05 before and after applying Bonferroni corrections respectively.}
    \label{tab:pvalues}
\end{table*}